\documentclass[sigconf]{acmart}

\usepackage{url}
\usepackage{float}
\usepackage{subfigure}
\usepackage{graphicx}
\AtBeginDocument{%
  \providecommand\BibTeX{{%
    \normalfont B\kern-0.5em{\scshape i\kern-0.25em b}\kern-0.8em\TeX}}}

\copyrightyear{2022} 
\acmYear{2022} 
\setcopyright{acmcopyright}\acmConference[CIKM '22]{Proceedings of the 31st ACM International Conference on Information and Knowledge Management}{October 17--21, 2022}{Atlanta, GA, USA}
\acmBooktitle{Proceedings of the 31st ACM International Conference on Information and Knowledge Management (CIKM '22), October 17--21, 2022, Atlanta, GA, USA}
\acmPrice{15.00}
\acmDOI{10.1145/3511808.3557470}
\acmISBN{978-1-4503-9236-5/22/10}

\begin{document}

\title{TFAD: A Decomposition Time Series Anomaly Detection Architecture with Time-Frequency Analysis}

% % \author{Paper ID: 2942}
% \author{Chaoli Zhang}
% \email{chaoli.zcl@alibaba-inc.com}
% \author{Tian Zhou}
% \email{tian.zt@alibaba-inc.com}
% \author{Qingsong Wen}
% \email{qingsong.wen@alibaba-inc.com}
% \affiliation{%
%   \institution{Institute for Clarity in Documentation}
%   \streetaddress{P.O. Box 1212}
%   \city{Dublin}
%   \state{Ohio}
%   \country{USA}
%   \postcode{43017-6221}
% }

\author{Chaoli Zhang}
\affiliation{%
  \institution{DAMO Academy, Alibaba Group}
  \city{Hangzhou}
  \state{}
  \country{China}
}
\email{chaoli.zcl@alibaba-inc.com}

\author{Tian Zhou}
\affiliation{%
  \institution{DAMO Academy, Alibaba Group}
  \city{Hangzhou}
  \state{}
  \country{China}
}
\email{tian.zt@alibaba-inc.com}

\author{Qingsong Wen}
\authornote{Corresponding author}
\affiliation{%
  \institution{DAMO Academy, Alibaba Group}
  \city{Bellevue}
  \state{}
  \country{USA}
}
\email{qingsong.wen@alibaba-inc.com}

\author{Liang Sun}
\affiliation{%
  \institution{DAMO Academy, Alibaba Group}
  \city{Bellevue}
  \state{}
  \country{USA}
}
\email{liang.sun@alibaba-inc.com}

% \author{
%     Chaoli Zhang\textsuperscript{\rm 1},
%     Tian Zhou\textsuperscript{\rm 1},
%     Qingsong Wen\textsuperscript{\rm 2},
%     Liang Sun\textsuperscript{\rm 2}
%     }
% \email{{chaoli.zcl, tian.zt, qingsong.wen, liang.sun}@alibaba-inc.com}
% \affiliation{
%     \institution{DAMO Academy, Alibaba Group}\\
%     \city{Hangzhou}
%     \country{China\textsuperscript{\rm 1}}
% }
% \affiliation{
%     \city{Bellevue}
%     \country{United States\textsuperscript{\rm 2}}
% }

\begin{abstract}

Time series anomaly detection is a challenging problem due to the complex temporal dependencies and the limited label data. Although some algorithms including both traditional and deep models have been proposed, most of them mainly focus on time-domain modeling, and do not fully utilize the information in the frequency domain of the time series data. In this paper, we propose a \textbf{T}ime-\textbf{F}requency analysis based time series \textbf{A}nomaly \textbf{D}etection model, or {\bf TFAD} for short, to exploit both time and frequency domains for performance improvement. Besides, we incorporate time series decomposition and data augmentation mechanisms in the designed time-frequency architecture to further boost the abilities of performance and interpretability. Empirical studies on widely used benchmark datasets show that our approach obtains state-of-the-art performance in univariate and multivariate time series anomaly detection tasks. Code is provided at https://github.com/DAMO-DI-ML/CIKM22-TFAD.

\end{abstract}



%%
%% Keywords. The author(s) should pick words that accurately describe
%% the work being presented. Separate the keywords with commas.

\maketitle

\section{Introduction}

%With the development of large systems (financial markets, Internet of Things systems, et al.) in reality, it is significant to monitor and detect anomalies in time series data to discover faults and avoid potential risks.
%Generally, an anomaly is an observation that deviates from normality, and its detection has been widely studied, including statistics, data mining, and machine learning \cite{ruff2021unifying}. 

With the rapid development of the Internet of Things (IoT) and other monitoring systems, there has been an enormous increase in time series data~\cite{robustts22,forecastFaloutsos20}. Thus, effectively monitoring and detecting anomalies or outliers on the time series data is crucial to discovering faults and avoiding potential risks in many real-world applications. Generally, an anomaly is an observation that deviates from normality. Anomaly detection has been studied widely in different disciplines, including statistics, data mining, and machine learning \cite{ruff2021unifying}, but how to perform it effectively on time series data is an active research topic and has received a lot of attentions recently~\cite{ren2019time,gao2020robusttad,zhang2021cloudrca,lai2021revisiting,tuli2022tranad,li2022learning,AnomalyKiTs_2022} due to the special properties of time series. 

% \cite{anomalyVLDB22,xu2021anomaly, }
% % 

Unlike ordinary tabular data, one distinguishing property of time series is the temporal dependencies. Usually, a point or a subsequence of time series is called an anomaly when compared to its corresponding ``context". Based on the relationship between the anomaly in time series and its context, we can define different types of anomalies, such as global point anomaly, seasonality anomaly, shapelet anomaly, etc. Thus, the first challenge in time series anomaly is how to model the relationship between a point/subsequence and its temporal context for different types of anomalies. Secondly, like other anomaly detection tasks, anomaly happens rarely, and there is usually limited labeled data for data-driven models. A possible solution is data augmentation, which is widely used in deep learning training~\cite{shorten2019survey}. Although some data augmentation methods have been proposed for time series data~\cite{wen2020time}, how to design and apply data augmentation in time series anomaly detection remains an open problem.

As a typical signal, the time series data can be analyzed not only from the time domain but also from the frequency domain~\cite{ts:textbook:1994}. Most of the existing methods, including conventional and deep methods, focus mainly on time-domain modeling and do not fully utilize the information in the frequency domain. The frequency domain can provide vital information for time series, such as seasonality~\cite{robustPeriod:2021}. In addition, it is much easier to detect in the frequency domain than in the time domain for some complex group anomalies and seasonality anomalies. Recently, there have been some attempts to model time series from the frequency domain~\cite{parhizkar2015sequences}, such as the data augmentation in the frequency domain~\cite{gao2020robusttad}. 
% we notice the difficulty to design a general frame single structure to detect anomalies both in time domain and frequency~\cite{parhizkar2015sequences}, 
Unfortunately, how to systematically and directly utilize the frequency domain and time domain information simultaneously in modeling time series anomaly detection is still not fully explored in the literature.

In this paper, to better detect various kinds of time series anomalies, we proposed a \textbf{T}ime-\textbf{F}requency domain analysis time series \textbf{A}nomaly \textbf{D}etection model, TFAD. It mainly contains two branches: the time-domain analysis branch and the frequency domain analysis branch. Specifically, with a well-designed window-based model structure, we implement a time series decomposition module to detect anomalies in different components with interference among different components reduced and a representation learning module with a neural network to gain richer sequence information. 
To deal with the challenges of insufficient anomaly data, we conduct data augmentation of TFAD in different views: normal data augmentation, abnormal data augmentation (not fully considered in existing works), time-domain data augmentation, and frequency domain data augmentation. 

% highlights of contributions:
% 1. Deep learning model from both time and frequency domains
% 2. Integrating time series decomposition into the deep learning framework
% 3. Systematic data augmentation methods for limited anomaly labels

In summary, our main contributions are listed as follows:
\begin{itemize}
\item We integrate the frequency domain analysis branch with the time domain analysis branch to identify the temporal information and improve detection performance. 
\item We combine the time series decomposition module with a concise neural representation network. With the help of time series decomposition, a simple temporal convolution neural network performs well. Moreover, it makes the model easy to be implemented, and the anomaly results of different components give insights into why it is abnormal.
\item Various data augmentation methods, besides normal data augmentation and time domain data augmentation, abnormal data augmentation and frequency domain data augmentation are also implemented to overcome the lack of anomaly data.
\end{itemize}

The rest of the paper is organized as follows. In Section~\ref{RelatedWork}, we review the related work. 
In Section~\ref{Preliminaries}, we briefly introduce the definitions of time series anomalies.
In Section~\ref{method}, we introduce our proposed TFAD algorithm, including motivations, architecture, and network design. In Section~\ref{experiments}, we evaluate our algorithm empirically on both univariate and multivariate time series datasets in comparison with other state-of-the-art algorithms. An ablation study is also performed to analyze different modules in the network. And we conclude our discussion in Section~\ref{Conclusion}.

\section{Related Work}\label{RelatedWork}

% \subsection{Anomaly Detection}
% 介绍已有的时序异常检测相关的文章和算法

% \subsection{Data Augmentation}
% \subsection{Time Series Decomposition}
% \subsection{Temporal Convolutional Networks}
% \subsection{Contrastive Learning}

% This section will briefly review the related architectures in time series anomaly detection. \cite{laptev2015generic} demonstrates a myth that there is an anomaly detection method that is optimal in all domains due to the complexity of anomalies in reality. \cite{lai2021revisiting} revisits the anomaly definitions and benchmarks in time series data. 

Both traditional and deep methods have been applied in time series anomaly detection tasks. 
A survey of traditional techniques for time series anomaly detection is in~\cite{gupta2013outlier}. 
Traditional techniques can be roughly classified as similarity-based~\cite{lane1997sequence, guan2016mapping}, window-based~\cite{gao2002hmms}, decomposition-based~\cite{gao2020robusttad, wen2019robuststl}, deviants detection based methods~\cite{muthukrishnan2004mining}, et al. 
With the rapid development of deep learning, a vast of deep anomaly detection methods emerge~\cite{kiran2018overview,hendrycks2018deep,ruff2021unifying}, including one-class type SVDD-based model~\cite{ruff2018deep}, reconstruction type GAN/VAE-based model~\cite{rezende2014stochastic}, et al. 
Deep methods are also applied in time series anomaly detection~\cite{shen2020timeseries, xu2021anomaly, zong2018deep}. 
% mainly credited to the powerful representation learning of deep neural network. 

% In another view, detection techniques vary depending on input data type, that is, three are different methods for uni-variate time series anomaly detection and multi-variate time series anomaly detection. 
% anomaly in time series data can be classified as point-wise and pattern-wise anomalies. 

% There is a vast sea of literature on time series anomaly detection. 

According to the input data type, there are methods designed for univariate time series, multivariate time series, or both. According to the access of labels, there are unsupervised, semi-supervised, and supervised time series anomaly detection methods. 
Online anomaly detection~\cite{chen2022adaptive, bock2022online} is also quite different from offline settings. 
For univariate time series anomaly detection, many traditional methods have been designed~\cite{ren2019time, siffer2017anomaly, xu2018unsupervised}. POT~\cite{siffer2017anomaly} proposes an approach based on Extreme Value Theory to detect outliers without the assumption of distribution. M-ELBO~\cite{xu2018unsupervised} designs a VAE-based algorithm on the Yahoo dataset. 
For multivariate time series anomaly detection, the development of representation learning stimulates blowout growth in the multivariate time series anomaly detection field. 
THOC~\cite{shen2020timeseries} proposes a temporal classification model for time series anomaly detection by capturing temporal dynamics in multi scales. 
It follows the one-class classification framework generally used~\cite{ruff2021unifying, ruff2018deep}. 
Anomaly Transformer~\cite{xu2021anomaly} designs unsupervised anomaly detection methods combining the transformer framework with a point-wise prior association by a mini-max strategy. 
Generative models, such as DAGMM~\cite{zong2018deep}, AnoGAN\cite{schlegl2017unsupervised}, and LSTM-VAE~\cite{park2018multimodal}, also form a mainstream research direction. 
Some surveys compare different kinds of anomaly detection methods in time series~\cite{freeman2021experimental, blazquez2021review}. 

The most related work of this paper is~\cite{carmona2021neural}, which introduces a window-based framework for anomaly detection in time series applied in unsupervised/supervised and univariate/multivariate settings. 
However, like most existing anomaly detection methods, it only works in the time domain. 
Although it works well for point-wise anomalies, it is usually hard to detect complex pattern anomalies, e.g., a sub-sequence time series. Instead, AutoAI-TS~\cite{AutoAI-TS} also takes the advantage of frequency domain analysis for period/seasonal forecasting.
In summary, our designed TFAD model contains two branches: the time-domain analysis branch and the frequency domain analysis branch, which is different from existing works. Besides, the time series decomposition module used in traditional methods is also implemented in our TFAD architecture. Furthermore, several well-designed data augmentation methods are also considered in TFAD to gain rich, reasonable, and reliable dataset. 

% In this paper, we mainly focus on anomaly detection in uni-variate time series in offline setting and leave the design for multivariate time series as future work.
% We combine a concise representation network, TCN (Temporal convolutional network)~\cite{bai2018empirical}, with the traditional time series analysis method to get a good performance. 
% TCN with few layers is not a strong representation network. Our design does not over-rely on the deep learning method. Instead, we are inspired by the power of traditional methods based on the situations. 

% More specifically, we follow the window-based framework in existing works to get semantics information but containing two branches: time domain analysis branch and frequency domain analysis branch. 

% Time series decomposition~\cite{huang1998empirical} has been widely used in traditional analysis and deep learning forecasting~\cite{wu2021autoformer}. We excavate its power when combined with the deep method in the anomaly detection field.  

% Correspondingly, to handle the common challenge in anomaly detection tasks that labeled data is usually expensive to get, we do data augmentation~\cite{wen2020time, oh2020time} not only in the time domain but also in the frequency domain. 

\section{Preliminaries}\label{Preliminaries}
% Time series anomaly detection has been widely studied. 
% Generally, the anomaly in non-sequential data is defined as an observation that deviates considerably from some concept of normality~\cite{ruff2021unifying}. 

In this Section, we will first give a brief view of the definitions of general anomalies. Anomalies appear in many situations. Anomalies in time series are a special type as the point without information of neighbor points means nothing. Such characteristics make time series analysis different from others. Thus, we will also show the definitions of time series or more specially, saying, sequential anomaly definitions. 

\subsection{General Anomaly Definitions}
In applications with non-sequential data, set $\mathcal{D}\in R$ as the data space and the normality follows distribution $\mathcal{N}^{+}$. Assuming $\mathcal{N}^{+}$ has a corresponding probability density function $p^{+}$, then the anomalies can be set as 
\begin{equation}
A=\{d\in \mathcal{D}\ | p^{+}(d) \leq \tau \}, \tau>0,
\end{equation}
where $\tau$ is the threshold of anomaly. 

Anomalies in non-sequential data can be mainly classified into three types: point anomaly, contextual anomaly, and group anomaly. 
A point anomaly is an individual data point that deviates from normality and is the most common case in anomaly detection. It can also be called a global anomaly. 
A context anomaly is also called a conditional anomaly. 
It is anomalous in a specific context. 
For instance, the temperature of 20 degrees is normal in most areas but abnormal in Antarctica.
A group or collective anomaly is a group of points that abnormal. 

% \begin{figure}[ht]
% \centering  
% \subfigure[Global Point Anomaly]{
% \label{sub1}
% \includegraphics[width=0.22\textwidth]{globalP.png}}
% \subfigure[Context Point Anomaly]{
% \label{sub2}
% \includegraphics[width=0.22\textwidth]{contP.png}}
% \caption{Examples of Point-wise Anomaly.}
% \label{point-anomalyClass}
% \end{figure}

\subsection{Sequential Anomaly Definitions}

% A time-series anomaly is a kind of anomaly in sequential data. 
% The definitions in non-sequential context can not sufficiently define the group and context anomalies in sequential data. 
% \cite{lai2021revisiting} refines sequential anomaly definitions. 

% To better model the temporal correlations among observations, behaviors of time series should be well-defined. 
Time series data is a sequence of data points $X=(x_0, x_1, x_2, \cdots, x_n)$ where $x_i$ is the point at timestamp $i$. 
More specially, time series can be formally defined by structural modeling~\cite{shumway2000time,lai2021revisiting} to include trend, seasonality and shapelets, as   
\begin{equation}
X =\sum_n \{ A\sin(2\pi\omega_n T) + B\cos(2\pi\omega_n T)\} + \tau(T),
\end{equation}
where $T=(1,2,3,\cdots, n)$ is a series of timestamps, $\omega_n$ is the frequency of wave $n$, $A, B$ are coefficients, the combination of sinusoidal wave represents the shapelets and seasonality, and $\tau(T)$ is trend component. 

\begin{figure}[t]
\centering  
% \caption{Examples of Point-wise Anomaly.}
\subfigure[shapelet anomaly]{
\label{sub2}
\includegraphics[width=0.15\textwidth]{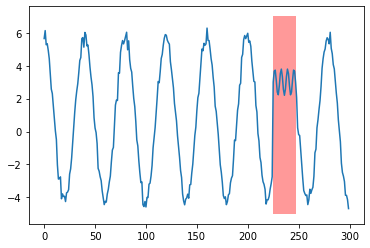}}
\subfigure[seasonal anomaly]{
\label{sub2}
\includegraphics[width=0.15\textwidth]{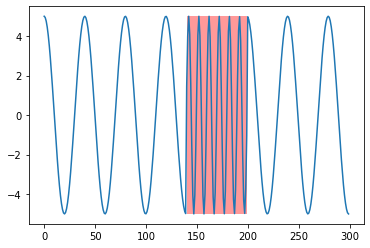}}
\subfigure[trend anomaly]{
\label{sub3}
\includegraphics[width=0.15\textwidth]{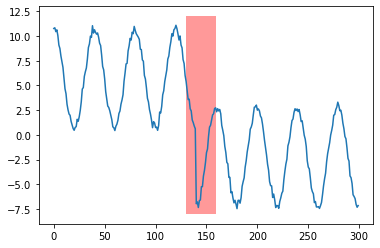}}
\subfigure[global point anomaly]{
\label{sub01}
\includegraphics[width=0.15\textwidth]{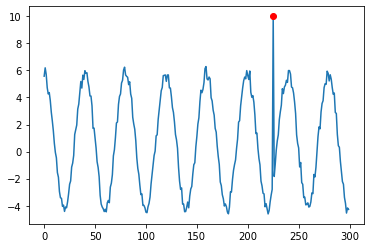}}
\subfigure[context point anomaly]{
\label{sub02}
\includegraphics[width=0.15\textwidth]{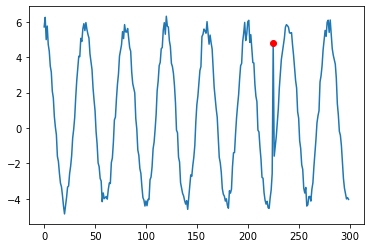}}
\caption{Anomalies in time series.}
\label{anomalyClass}
\end{figure}

The definitions in a non-sequential context can not sufficiently define the group and context anomalies in sequential data. Instead, sequential anomalies can be classified as point anomaly (global point anomaly and context point anomaly) and pattern anomaly (shapelet anomaly, seasonal anomaly, and trend anomaly)~\cite{lai2021revisiting}, as shown in Figure~\ref{anomalyClass}.

Formally, point-wise anomalies can be defined as ${|x_t - \hat{x_t}|}>{\sigma}$ where $\hat{x_t}$ is the expected value and $\sigma$ is threshold.
Shapelet anomaly can be defined as $s(\rho(\cdot), \hat{\rho}(\cdot)) > \sigma,$ where function $s$ measures the difference between two subsequences, $\hat{\rho}$ is the expected shapelet and $\sigma$ is a threshold.
Seasonal anomaly refers to subsequence with abnormal seasonality and it is defined as $s(\omega, \hat{\omega}) > \sigma$ where $\hat{\omega}$ is expected seasonality. 
Trend anomaly is a subsequence whose trend alters the trend of $X$. It can be defined as $s(\tau(\cdot), \hat{\tau}(\cdot))>\sigma$ where $\hat{\tau}$ is the expected trend of subsequence.

\section{Proposed TFAD Algorithm}\label{method}
\subsection{Motivation: Time-Frequency Analysis}
In this part, we will provide the design motivation of TFAD algorithm through the uncertainty principle of time-frequency analysis and demonstrate its effectiveness in time series anomaly detection. 

\subsubsection{Uncertainty Principle for Time Series Representation in Time and Frequency Domains}

The uncertainty principle expresses a fundamental relationship between the standard deviation of a continuous function and the standard deviation of its Fourier transformation~\cite{cohen1995time}. Let the input signal $s(t)$ have a spectrum $S(\omega)$ as
\begin{equation}
S(\omega)=\frac{1}{\sqrt{2 \pi}} \int_{-\infty}^{\infty} s(t) e^{-j \omega t} d t.
\end{equation}
The standard deviations of the time and frequency density functions, $\sigma_{t}$ and $\sigma_{\omega}$, are defined as the parameters that describe the broadness of the signal in time and frequency domains, respectively
\begin{equation}
\sigma_{t}^{2}=\int(t-<t>)^{2}|s(t)|^{2} d t, ~~
\sigma_{\omega}^{2}=\int(\omega-<\omega>)^{2}|S(\omega)|^{2} d \omega.
\end{equation}
% \begin{equation}
% \begin{aligned}
% \sigma_{t}^{2}&=\int(t-<t>)^{2}|s(t)|^{2} d t, \\
% \sigma_{\omega}^{2}&=\int(\omega-<\omega>)^{2}|S(\omega)|^{2} d \omega.
% \end{aligned}
% \end{equation}
Then we have the following result based on Schwarz inequality
\begin{equation}
\begin{aligned}
\sigma_{t}^{2} \sigma_{\omega}^{2} &=\int|t s(t)|^{2} d t \times \int\left|s^{\prime}(t)\right|^{2} d t  \\
& \geq\left|\int t s^{\star}(t) s^{\prime}(t) d t\right|^{2}=\left|-\frac{1}{2}+j \operatorname{Cov}_{t \omega}\right|^{2}=\frac{1}{4}+\operatorname{Cov}_{t \omega}^{2}.
\end{aligned}
\end{equation}
% Hence 
% $$
% \sigma_{t}^{2} \sigma_{\omega}^{2} \geq\left|\int t s^{\star}(t) s^{\prime}(t) d t\right|^{2}=\left|-\frac{1}{2}+j \operatorname{Cov}_{t \omega}\right|^{2}=\frac{1}{4}+\operatorname{Cov}_{t \omega}^{2}
% $$
When we view the uncertainty principle from time series anomaly detection, it means that when one structure has a good catch of point-wise anomaly in the time domain, it would have a less sensitive detection power in the frequency domain, and vice versa. Designing a model with a single structure that can detect the time and frequency simultaneously is difficult. A better approach could be using different structures to detect them and merge them as a whole model.

\begin{figure}[t]
\centering  
\subfigure[Point-wise anomaly example]{
\label{PDt}
\includegraphics[width=0.151\textwidth]{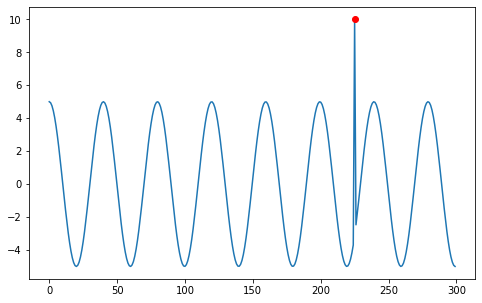}}
\subfigure[Diff in time domain w/o point-wise anomaly]{
\label{PDdt}
\includegraphics[width=0.151\textwidth]{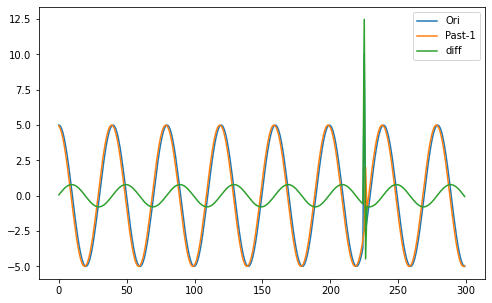}}
\subfigure[Diff in frequency domain w/o point-wise anomaly]{
\label{PDdf}
\includegraphics[width=0.151\textwidth]{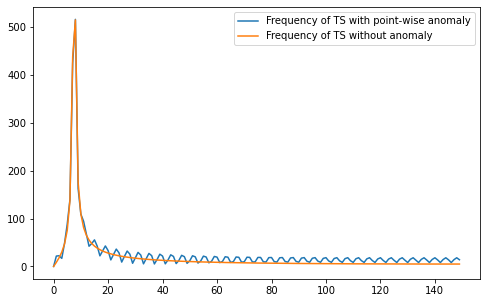}}
\caption{Examples of point-wise anomaly in time and frequency domain.}
\label{exPoint}
\end{figure}

\begin{figure}[t]
\centering  
\subfigure[Seasonality anomaly example.]{
\label{SDt}
\includegraphics[width=0.151\textwidth]{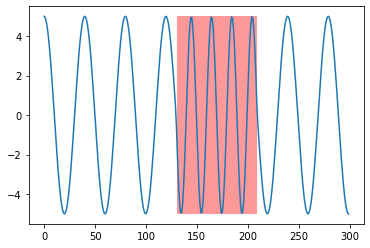}}
\subfigure[Diff in time domain w/o seasonality anomaly]{
\label{SDdt}
\includegraphics[width=0.151\textwidth]{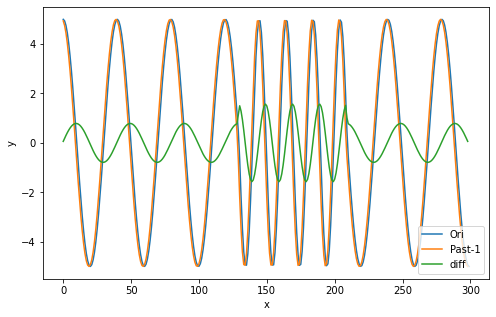}}
\subfigure[Diff in frequency domain w/o seasonality anomaly]{
\label{SDdf}
\includegraphics[width=0.151\textwidth]{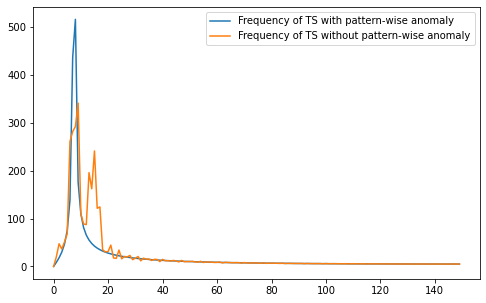}}
\caption{Examples of pattern-wise anomaly in time and frequency domain.}
\label{exPattern}
\end{figure}

\subsubsection{Understanding the Limitations of Single Domain Analysis}\label{SDexample}

This subsection will show the significant difference in point-wise anomaly detection and pattern-wise anomaly detection intuitively by two simple examples and demonstrate the limitations of detecting anomalies only in the time or frequency domain. The illustration examples are plotted in Figure~\ref{exPoint} and Figure~\ref{exPattern}, where
%For simplicity, we assume the expected value of anomalies can be gained ideally, for example, by the forecasting method. 
Figure~\ref{exPoint} shows the differences between the detection of point-wise anomaly in time domain (fig~\ref{PDdt}) and frequency domain (fig~\ref{PDdf}) by a global point anomaly example (fig~\ref{PDt}), and 
Figure~\ref{exPattern} shows the differences between the detection of pattern wise anomaly in time domain (fig~\ref{SDdt}) and frequency domain (fig~\ref{SDdf}) by a seasonality anomaly example (fig~\ref{SDt}).

Generally, anomaly detection in time domain is done by comparing the point with its past neighbors. For simplicity, assume the time step is one between two adjacent points.
We show the analysis in time domain intuitively by computing the difference (noted as diff) between the original point (noted as Ori) and the point before it (noted as Past-1). 
Figure~\ref{PDdt} and figure~\ref{SDdt} show the different results in time domain of the point-wise anomaly example and pattern-wise anomaly example, respectively. 
It can be seen that point anomaly is easier to be detected than seasonality anomaly in time domain analysis. Detecting seasonality anomalies only through time-domain analysis is not easy.

% Figure~\ref{exFreq} shows the differences between the detection of point wise anomaly and the detection pattern-wise anomaly in \textbf{frequency} domain by two typical examples (fig~\ref{PDsub1}, fig~\ref{SDsub1}). 

For anomaly detection in the frequency domain, we compare the time series results with/without anomalies after time to frequency transform (by Fourier transform).
Figure~\ref{PDdf} shows results of the time series in frequency domain with and without point-wise anomalies. The difference is subtle and disperses in many channels, indicating it is hard to detect point-wise anomaly only with time domain analysis.
Figure~\ref{SDdf} shows the results of time series in frequency domain with and without seasonality-wise anomaly. Unlike the situation in point-wise anomalies, the difference is quite evident as different numbers of peaks are shown.
Thus, seasonality anomaly is easier to detect than point anomaly in frequency domain analysis. It is impractical to detect point-wise anomalies only by frequency domain analysis.

% From Fig~\ref{PDsub1}\textasciitilde Fig\ref{PDsub6}, it can be seen that the global point anomaly is evident and easy to be detected in the time domain but subtle in many frequency channels and hard to be detected in the frequency domain. 
% On the contrary, the seasonality anomaly is easier to be detected in the frequency domain by fewer channels (evident drop in frequency around eight and evident rise in frequency around 30) than in the time domain, which is shown in Fig~\ref{SDsub1}\textasciitilde Fig~\ref{SDsub6}. 
% More analysis, in theory, is in Section~\ref{TheoryAna}. 

% \subsection{Intuition of Model Design}
% \subsection{Motivation: Time-Frequency Analysis}

\subsection{Motivation: Data Augmentation and Decomposition}
Besides the time-frequency analysis, we also consider data augmentation and decomposition to further improve the performance of time series anomaly detection.

% The most common way is to split trends from time series. 

% There are different kinds of methods to make time-series decomposition, for example, moving averages, classical decomposition methods such as additive decomposition and multiplicative decomposition, X11 decomposition method~\cite{dagum2016seasonal}, seat decomposition method~\cite{dagum2016seasonal}, and STL decomposition method~\cite{robert1990stl}. 

% Time series decomposition makes it possible to analyze trend and seasonality separately. 

\subsubsection{Time Series Data Augmentation}

The performance of machine learning usually relies on many training data. 
However, in reality, labeled data is usually limited. Data augmentation~\cite{wen2020time} contributes a lot to help mitigate these challenges.
Unlike most existing works where augmented data should follow the original data distribution, we consider two kinds of data augmentation methods for the anomaly detection task:
Data augmentation for normal data and anomaly data.
In the anomaly detection task, anomalies are samples different from normal data. There are usually various kinds of anomalies. 
Thus, when anomaly data is augmented, there is no need to create anomalies identical to real anomalies in the dataset, which is also impractical. Diverse augmented anomalies generally contribute to the robustness of models.

\subsubsection{Time Series Decomposition}

Generally speaking, time-series data often exhibit various patterns, and it is usually helpful to split a time series into main components. 
It is a powerful technology for analyzing complex time series widely adopted in time series anomaly detection~\cite{hochenbaum2017automatic,gao2020robusttad,zhang2021cloudrca} and forecasting~\cite{FedFormer,xu2021autoformer,quatformer22}.
Furthermore, with the help of time series decomposition, simple temporal convolution neural networks can bring desirable performance (this will be discussed later). It makes the model easy to be implemented and provides insights based on anomaly results of different components.

\subsection{High Level Architecture of TFAD}
\begin{figure}[t] 
\centering 
\includegraphics[scale=0.2]{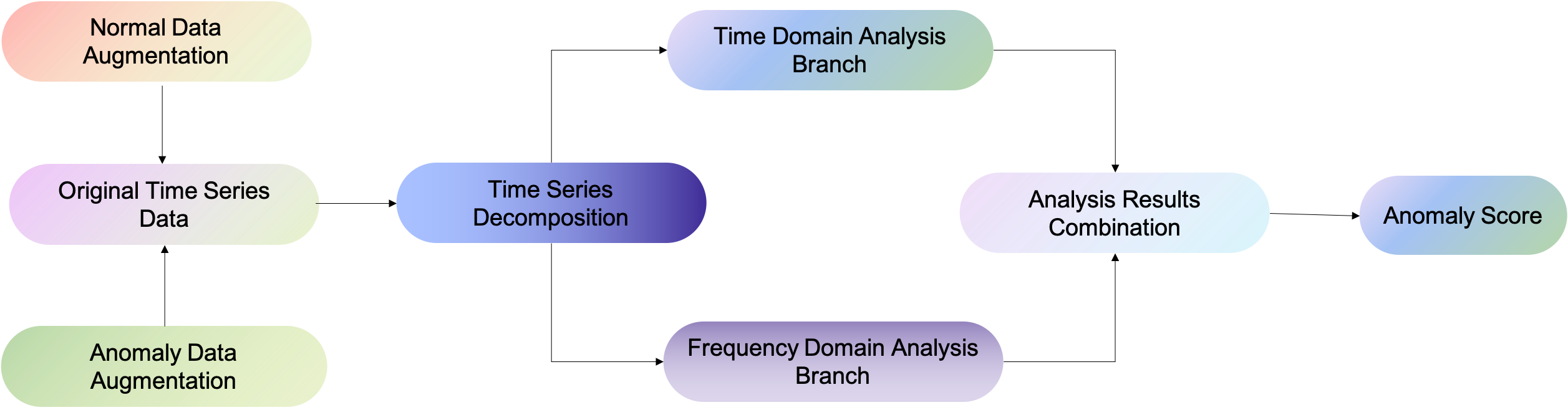} 
\caption{High-level diagram of TFAD architecture.} 
\label{overview} 
\end{figure}

Following the aforementioned motivations, the high level architecture of our designed TFAD algorithm is summarized in Fig~\ref{overview}. Our method consists of two main branches, the time-domain analysis branch and the frequency-domain analysis branch. Besides that, both normal data augmentation module and anomaly data augmentation module are designed to increase the robustness of our method. Furthermore, the time series decomposition module is adopted to better detect anomalies in different components and provide insights into the explanation of anomalies. 

% we proposed our TFAD algorithm: a time series decomposition anomaly detection architecture with Time-Frequency analysis. 
% The overview architecture of TFAD is shown in Fig~\ref{overview}. 
% The following modules are contained: normal data augmentation module, anomaly data augmentation module, time series decomposition module, time domain analysis module, and frequency domain analysis module. 

% \subsection{Overview Architecture of TFAD}

% With frequency domain analysis added, the property of time frequency domain can be effectively combined. The robustness of TFAD is also improved. The intuition is that, with complex anomaly types, only time-domain analysis or only frequency domain analysis results in over-fitting in special anomalies. Time-domain analysis can not work well in seasonal anomalies, which leads to over-fitting in the training set and thus can not get good and reliable detection in seasonality anomalies. Frequency domain analysis can not work well in point anomalies and similar problem happens. 

\subsection{Network Design of TFAD}
\begin{figure*}[ht] 
\centering 
\includegraphics[scale=0.25]{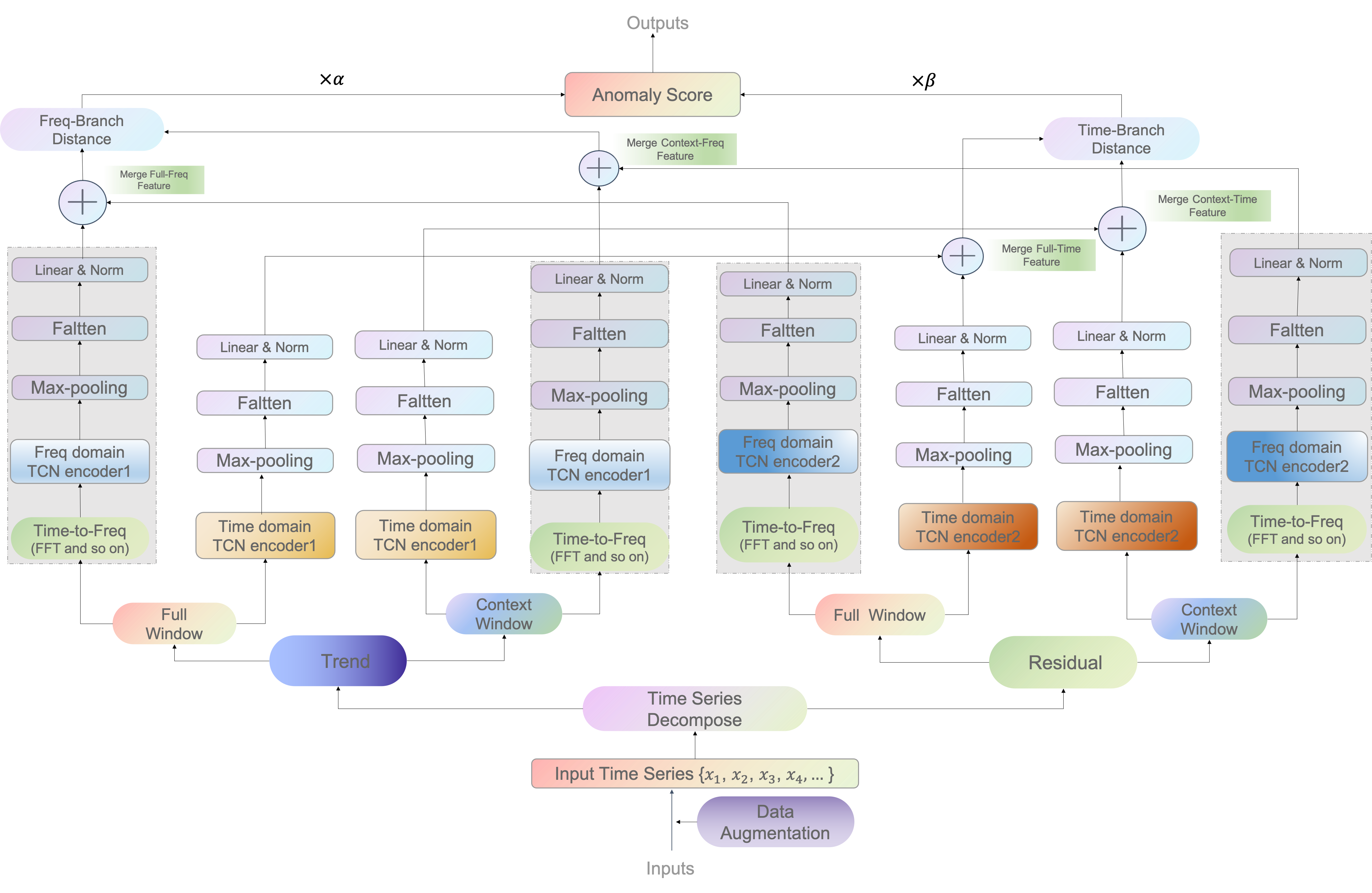} 
\caption{Network structure of the proposed TFAD algorithm. The shaded blocks indicate the frequency domain branch.} 
\label{network} 
\end{figure*}

In this section, we provide the detailed design of the TFAD algorithm. The whole network structure of TFAD is plotted in Fig.~\ref{network}, where each module will be elaborated in the following parts.

% In this section, we will show the details of the modules in TFAD.

% \subsubsection{Normal data augmentation}\label{normAug}
\subsubsection{Data Augmentation Module}\label{normAug}
As discussed before, we consider both normal and anomaly data augmentation.

% In normal data augmentation module, besides augmenting similar data to normal ones, we also gain \emph{more normal} data with noise reduction methods (e.g., Robust STL, et al). 
% The anomaly data augmentation module is conducive to generating a diversity of anomalies, that is, including anomalies not entirely consistent with real anomalies in datasets. 

% More details can be found in Section~\ref{normAug} and Section~\ref{abnormAug}, respectively.

\begin{figure}[t] 
\centering 
\includegraphics[scale=0.3]{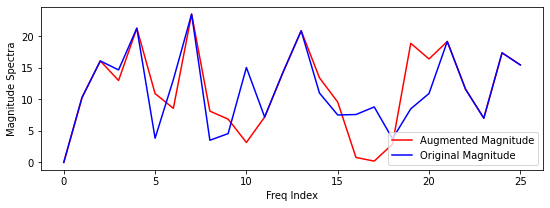} 
\includegraphics[scale=0.3]{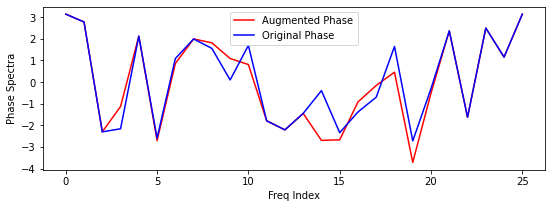} 
\includegraphics[scale=0.3]{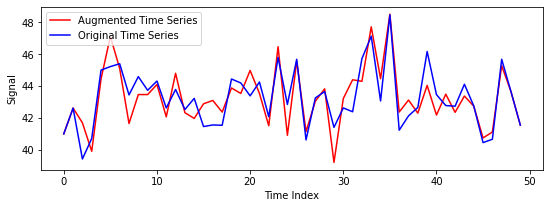} 
\caption{Norm data augmentation by frequency domain changes.} 
\label{NormAug} 
\end{figure}

For normal data augmentation, firstly, we generate data with low noise, which is \emph{more normal}. Robust STL~\cite{wen2019robuststl} is a desirable option to get trend and seasonal information of time series data, and the residual, which is in some way noise, can be ignored. 
% Besides that, we also reduce noises of original data to gain more \emph{normal} data. 
Such \emph{more normal} data keep the innate character of normal patterns and have larger differences with anomalies. It is easier to distinguish \emph{more normal} data from anomalies and helps our model learn the intrinsic quality of normal data. 
To create diverse normal data, we transfer time series to the frequency domain by Fourier transform and make small changes in both the imaginary and the real parts to gain new data. In this way, diverse data is augmented. An example is shown in Figure~\ref{NormAug}.

% \subsubsection{Anomaly data augmentation}\label{abnormAug}
Besides the typical anomalies in the data set for anomaly data augmentation, other possible anomalies should also be considered for anomaly data augmentation. 
It helps to detect new anomalies and contributes to the robustness of our method. 
Similar results also appear in the computer vision field~\cite{ruff2020rethinking}, which demonstrates that relatively few random outlier exposure images help to yield state-of-the-art detection performance. 
Specifically, we consider several data augmentation methods. 
Point scale modification in the time domain is adopted for point change anomaly, which is the most common type.
For context anomalies, point/short sequence exchange and a mix-up between two different time series are considered. 
For anomalies in seasonal and some other complicated anomalies, several different data augmentation methods in the frequency domain are used to generate various sequence anomalies. 

\subsubsection{Decomposition Module}
There are different kinds of methods to make time-series decomposition, for example, moving averages, classical decomposition methods such as additive decomposition and multiplicative decomposition, X11 decomposition method~\cite{dagum2016seasonal}, seat decomposition method~\cite{dagum2016seasonal}, and STL decomposition method~\cite{robert1990stl}. 
In this paper, Hodrick–Prescott (HP) filter~\cite{hodrick1997postwar} is adopted for time series decomposition since it is easy to implement and works well in the real world.

% Denote time series $y_t, t=1,2, \dots, T$ contains a trend component ${\tau}_t$, a residual part ${\epsilon}_t$ and a possible seasonal component $c_t$. That is, $y_t = \tau_t + c_t + \epsilon_t$. Then, in HP filter, the trend component can be obtained by solving the following minimization problem
Denote time series $y_t$, $t=\{1,2, \dots, T\}$ contains a trend component ${\tau}_t$, a residual part ${\epsilon}_t$. That is, $y_t = \tau_t + \epsilon_t$. Then, in HP filter, the trend component can be obtained by solving the following minimization problem
\begin{equation}
{\min}_{\tau} \left( \sum_{t=1}^{T}(y_t - \tau_t)^2 + \lambda\sum_{t=2}^{T-1}{[(\tau_{t+1} - \tau_{t}) - (\tau_t - \tau_{t-1})]}^2 \right),
\end{equation}
where the multiplier $\lambda$ is the parameter that adjusts the sensitivity of the trend to fluctuation and can be adjusted according to the frequency of observations~\cite{ravn2002adjusting}. After decomposition, both trend and residual components are utilized to improve performance since different anomalies may appear in different components.

We mainly consider the decomposition method for univariate time series in TFAD. 
For multivariate time series, we decompose each time series sequence separately. 
It may not be the best method for multivariate time series decomposition. However, significant improvement is gained, and we will leave a special design for multivariate time series as future work.

\subsubsection{Window Splitting Module}

% \subsection{Network Structure}{\label{NetworkStruc}}
 
\begin{figure}[t] 
\centering 
\includegraphics[scale=0.4]{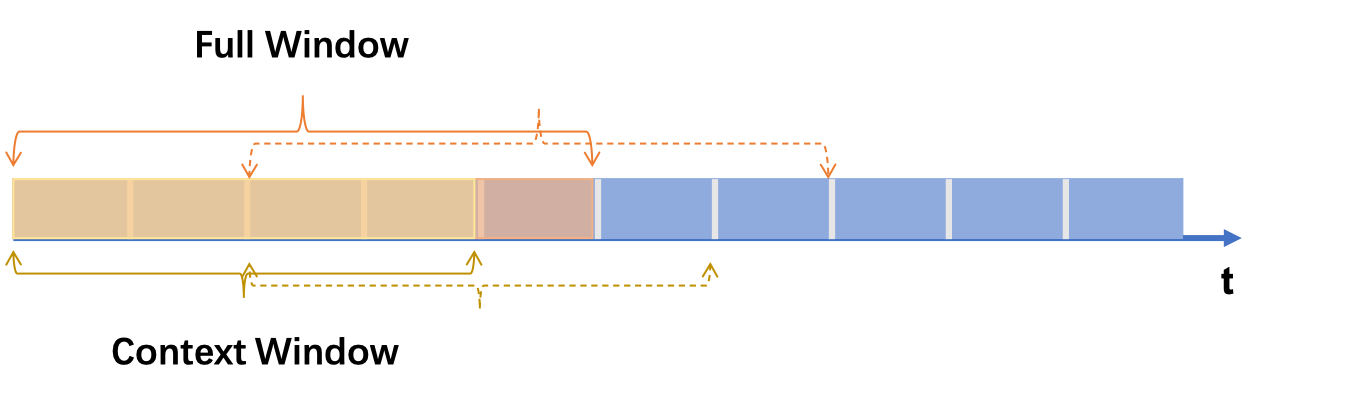} 
\caption{Window splitting.} 
\label{window} 
\end{figure}

 In the time series anomaly detection task, temporal correlations among observations are significant, and sequence anomalies are usually harder to be detected than point anomalies. 
 Furthermore, even point anomaly is hard to be detected without temporal correlations. 
 Therefore, we adopt time series window to better gain sequence-wise/temporal correlation information~\cite{carmona2021neural, tuli2022tranad}. Specifically, a full time series window and a context time series window are set to detect anomalies in a suspect sequence, where the full window consists of a context window and a suspect window, as illustrated in Figure~\ref{window}. The assumption is that suppose the context window is normal. If the pattern of the full window is consistent with the pattern of the context window, then there is no anomaly in the suspect window. 
 If there is an anomaly in the suspect window, the pattern of the full window will not be consistent with the context window. With sliding windows, the label of each time point can be known and more details are in Section~\ref{AS}.

\subsubsection{Time and Frequency Branches}\label{FFT}
 In this section, the time and frequency branches will be discussed. 
 As shown in Fig~\ref{network}, the original time series will be first decomposed into trend and residual components. 
 For each component, we set the full window sequence and context window sequence with the aforementioned window splitting. After that, time-domain representation learning and frequency-domain representation learning for each window sequence will be done to gain rich information on sequences. After that, the distance between the context window and the full window would be measured to calculate the anomaly score.

For the representation, most of the classical distances, such as Cosine distance, dynamic time warping (DTW) distance, are too susceptible to the length of time series to be used here. 
Instead, neural networks are widely used to gain the representation of complex samples in many tasks due to the power of representation ability. Thus, we utilize a neural representation network to overcome the above shortcoming of classical methods. 
Specifically, the temporal convolutional network (TCN)~\cite{bai2018empirical} is a simple and powerful architecture. Therefore, we adopt the TCN design as our representation network. 

TFAD first transforms the time series from time-domain to frequency-domain for the frequency branch by discrete Fourier transform (DFT).
For time series $\{x_n\}: = x_0, x_1, \dots, x_{N-1}$, its discrete Fourier transform $\{X_k\}:=X_0, X_1, \dots, X_{N-1}$ is defined as
\begin{equation}
    X_k = \sum_{n=0}^{N-1}x_n\cdot e^{-\frac{i2\pi}{N}kn} \\
= \sum_{n=0}^{N-1}x_n\cdot\left[ cos(\frac{2\pi}{N}kn) - i\cdot sin(\frac{2\pi}{N}kn)\right].
\end{equation}
The results of DFT contain real-part (denoted as $Re$) and imaginary-part (denoted as $Im$). To better get the information of time series in different locations, TFAD intersects $Re$ and $Im$ and gets $\{X_k\} = F_{\{x_n\}} = \{Re_1, Im_1, Re_2, Im_2, \dots, Re_n, Im_n\}$.
After obtaining DFT results, representation learning is done by TCN similarly to the time branch. 
 
There are different transform methods for time-frequency translation besides DFT, such as continuous Fourier Transform (CFT), short-time Fourier Transform (STFT), and wavelet transform. DFT is more suitable for our situation:
the time-series data is usually discrete.
As the length of the anomaly sequence is unknown, the window length of STFT is hard to set.
Wavelet transform is considered to contain time and frequency information simultaneously, but evaluation shows that it does not gain a good performance as DFT.
The main reason may be that, with the time-domain branch added already, a DFT with frequency information can gain more marginal utility than a wavelet.

\subsubsection{Anomaly Score Module}\label{AS}
  
 Comparison between full window sequence and context window sequence is used to set anomaly score, which is a widely used metric for the \emph{degree of anomalousness}~\cite{ruff2021unifying}. Here, the cosine similarity between full window sequence and context window sequence is set as anomaly score. Specifically, denote anomaly score as $AS$, then 
 \begin{equation}
AS = \mathcal{F} \{dis(RV_{treT}, RV_{resT}, RV_{treF}, RV_{resF})\},
 \end{equation}
where $RV_{treT}$, $RV_{resT}$ are representation results of trend component and residual component in \emph{time} domain respectively, $RV_{treF}, RV_{resF}$ are representation results of trend component and residual component in \emph{frequency} domain respectively, and the distance function $dis()$ is the cosine similarity. 
 The higher the anomaly score is, the higher the dissimilarity is, which means the suspect window is more likely to be abnormal. We can decide whether the suspect window is abnormal with a threshold for anomaly score.
 Thus, suspect windows can be labeled with an anomaly or not. 
 However, we also want to know what label should be set for every time point. 
 Therefore, a voting strategy is adopted. With sliding full and context windows, the suspect window is also sliding. Every point belongs to several suspect windows, and if more than half of them are labeled as an anomaly, the point will be set as an anomaly.

\section{Experiments}\label{experiments}

% In this section, we evaluate the performance of our method on commonly used benchmark data sets. To show the advantage of TFAD more directly, we evaluate our algorithm mainly on univariate time-series datasets. It is intuitive to get the trend and cycle part from univariate TS. We leave the application in multivariates as future work. 

This section studies the proposed TFAD model empirically compared to other state-of-the-art time series anomaly detection algorithms on both univariate and multivariate time series benchmark datasets. We also investigate how each component in TFAD contributes to the final accurate detection by ablated studies and discuss the insights.

\subsection{Baselines, Datasets, Metrics, and Evaluation}
\subsubsection{Baselines.} For univariate time series anomaly detection, we compare our method with the state-of-the-art algorithms, including SPOT, DSPOT~\cite{siffer2017anomaly}, DONUT~\cite{xu2018unsupervised}, SR, SR-DNN, SR-CNN~\cite{ren2019time}, and NCAD~\cite{carmona2021neural}. For multivariate time series anomaly detection, we compare recent deep neural network models like
% Many deep neural network models have recently emerged for multivariate time series anomaly detection. Most of them, such as, 
AnoGAN~\cite{schlegl2017unsupervised}, DeepSVDD~\cite{ruff2018deep}, DAGMM~\cite{zong2018deep}, LSTM-VAE~\cite{park2018multimodal}, MSCRED~\cite{zhang2019deep}, OmniAnomaly~\cite{su2019robust}, MTAD-GAT~\cite{zhao2020multivariate}, THOC~\cite{shen2020timeseries}. Note that some baselines above are not designed for temporal data, but they are extended for time series data with fixed lengths by sliding windows.

% but with , the time series inputs with fixed lengths can be considered as general examples. 

\subsubsection{Datasets.} We adopt the widely-adopted univariate and multivariate datasets for time series anomaly detection as follows: 
\begin{itemize}
\item KPI~\cite{kpidata} is a univariate time series dataset released in the AIOPS anomaly detection competition. It contains dozens of KPI curves with labeled anomaly points. The points are collected every 1 minute or 5 minutes from Internet Companies, for instance, Sogou, Tencent, eBay, etc. 

\item Yahoo~\cite{yahoolink} is a univariate time series dataset for time series anomaly detection released by Yahoo research. Part of the dataset is synthetic, where the anomalies are algorithmically generated. Part of it is real traffic data to Yahoo services, where the anomalies are labeled manually by editors. 
    
\item SMAP and MSL~\cite{hundman2018detecting} are two multivariate time series datasets published by NASA. SMAP and MSL have 55 and 27 unique telemetry channels, respectively, that is, 55 and 27 dimensions time series. More specifically, anomaly sequences in SMAP are composed of $62\%$ point anomalies and $38\%$ contextual anomalies, while MSL consists of $53\%$ point anomalies and $47\%$ contextual anomalies. 

\end{itemize}
   
The summary of these datasets is shown in Table~\ref{tab:summary}.

\begin {table}
\caption {Summary of adopted datasets.} \label{tab:summary} 
\begin{center}
\begin{tabular}{c|c|c|c}
\hline
data set & $\#$Curves/Dims & $\#$Points & $\%$Anomaly \\
\hline
KPI & 58 & 5922913 & 2.26 \\
Yahoo & 367 & 572966 & 0.68 \\
SMAP & 55 & 429735 & 12.8\\
MSL  & 27 & 66709 &  10.5\\
\hline
\end{tabular}
\end{center}
\end{table}

\subsubsection{Metrics.} Point adjusted F1 score~\cite{shen2020timeseries, audibert2020usad, su2019robust} is the widely used metric in the time series anomaly detection task. In this metric, if one point is detected as an anomaly in a segment, the whole anomaly segment will be considered as detected. This metric fits well with real-world situations as, in most cases, the anomaly event affects more than one time point. For such an abnormal event, a single anomaly alarm is enough. Note that several other F1-type metrics~\cite{kim2021towards, garg2021evaluation, jacob2020exathlon, hwang2022you} have been proposed to provide a more precise evaluation of abnormal event detection. Either the first alarm of group anomalies is set with higher importance, or the proportion of alarms is evaluated. However, in this paper, our primary aim is not to discuss which metric is the best as different metrics are applied in different situations. Thus, we adopt the widely used point-adjusted F1 score as our metric and leave evaluations on more different metrics for future work.
% and precision will also be shown to evaluate our design in special cases. 

\subsubsection{Evaluation Details.} We follow the common setting as in~\cite{ren2019time, carmona2021neural} for better comparisons. Specifically, we split each dataset into the train part, validation part, and test part to choose models. For the Yahoo dataset, we split 50\% as test data, 30\% as train data, and 20\% as validation data.
The original KPI dataset contains train and test data, and we set 30\% of the training data as validation data. 
For the evaluation of the KPI dataset, we apply both supervised setting (sup.) where all labeled data are utilized and unsupervised setting (un.) where the label information is not utilized.
Every setting has been run ten times, and the mean and variance are reported.

\subsection{Performance Comparisons}
The performance comparisons of different baseline algorithms and our TFAD are summarized in Table~\ref{tab:unires} and Table~\ref{tab:mulres} for univariate and multivariate time series anomaly detection tasks, respectively.

For the univariate time series anomaly detection in Table~\ref{tab:unires}, it can be seen that deep learning methods usually bring better performance than the conventional methods like SPOT and DSPOT, due to their strong representation abilities.  
Note that both NCAD~\cite{carmona2021neural} and our TFAD introduce data augmentation, and the randomness of augmented data brings in fluctuation in the F1 score. The variance of TFAD is significantly lower than NCAD in most cases, which indicates our TFAD method would be more robust and stable in practical systems. 
In summary, our TFAD algorithm produces comparable performance to the best NCAD algorithm on the Yahoo dataset, and significantly outperforms all other competing algorithms on the KPI dataset.

\begin {table}[t]
\caption {F1 score of anomaly detection on \textit{univariate} time series datasets. The best results are highlighted.} \label{tab:unires} 
\begin{center}
\begin{tabular}{c|c|c|c}
\hline
% Model & Yahoo (semi-sup) & KPI (semi-sup)  & KPI (sup) \\
Model & Yahoo (un.) & KPI (un.)  & KPI (sup.) \\
\hline
SPOT & 33.8 & 21.7 & -- \\
DSPOT & 31.6 & 52.1 & -- \\
DONUT & 2.6 & 34.7 & -- \\
SR  & 56.3 & 62.2 & -- \\
SR-CNN & 65.2 & 77.1 & -- \\
SR-DNN & -- & -- & 81.1 \\
% NCAD & $80.531\pm 1.42$ & $76.64\pm 0.89$ & $79.2\pm 0.92$ \\
NCAD & $\textbf{81.16}\pm 1.43$ & $76.64\pm 0.89$ & $79.20\pm 0.92$ \\
\hline
% TFAD & $\textbf{81.127}\pm \textbf{0.52}$ & $\textbf{79.8}\pm \textbf{0.74}$ & $\textbf{82.1}\pm \textbf{0.42}$ \\
\textbf{TFAD} & ${81.13}\pm \textbf{0.52}$ & $\textbf{79.80}\pm \textbf{0.74}$ & $\textbf{82.10}\pm \textbf{0.42}$ \\
\hline
\end{tabular}
\end{center}
\end{table}

\begin{table}[t]
\caption {F1 score of anomaly detection on \textit{multivariate} time series datasets. The best results are highlighted. } \label{tab:mulres} 
\begin{center}
\begin{tabular}{c|c|c}
\hline
Model & SMAP (un.) & MSL (un.)  \\
\hline
AnoGAN & 74.59 & 86.39  \\
DeepSVDD & 71.71 & 88.12  \\
DAGMM & 82.04 & 86.08  \\
LSTM-VAE  & 75.73 & 73.79  \\
MSCRED & 77.45 & 85.97  \\
OmniAnomaly & 84.34 & 89.89  \\
MTAD-GAT & 90.13 & 90.84 \\
THOC & 95.18 & 93.67 \\
NCAD & $94.45\pm \textbf{0.68}$ & $95.60\pm 0.59$ \\
\hline
% TFAD & $\textbf{96.324}\pm 1.57$ & $\textbf{96.409}\pm \textbf{0.34}$ \\
\textbf{TFAD} & $\textbf{96.32}\pm 1.57$ & $\textbf{96.41}\pm \textbf{0.34}$ \\
\hline
\end{tabular}
\end{center}
\end{table}

% FreqBran & ${13.81}\pm {3.08}$ & ${35.97}\pm {4.40}$ & ${19.59}\pm {2.81}$ \\
\begin {table*}[t] %跨栏加*
% \caption {Results of ablation study on KPI data.} \label{tab:ablation} 
\caption {Ablation studies of the proposed TFAD on KPI dataset. Denote the time series decomposition module as \textbf{Dec}, the norm data augmentation module as \textbf{NorAug}, the time domain anomaly data augmentation module as \textbf{TimeAnAug}, the frequency domain anomaly data augmentation module as \textbf{FreqAnAug}, and the frequency domain analysis module as \textbf{FreqBran}. The proposed TFAD algorithm combines all these modules. The best results are highlighted.} \label{tab:ablation} 
\begin{center}
\begin{tabular}{c|c|c|c|c|c|c|c|c|c}
\hline
case & TCN & Dec & NorAug & TimeAnAug & FreqAnAug & FreqBran & Precision & Recall &  F1 score\\
\hline
Freq Branch &  & & & & & $\checkmark$ & ${13.81}\pm {3.08}$ & ${35.97}\pm {4.40}$ & ${19.59}\pm {2.81}$\\
Time Branch & $\checkmark$ & & & & & & $44.888\pm0.095$ & $57.227\pm0.0381$ & $50.312\pm 0.0569$\\
(a) & $\checkmark$ & $\checkmark$ & & & & &$57.557\pm5.374$ & $81.111\pm5.339$& $66.968\pm2.58$\\
(b) & $\checkmark$ & $\checkmark$ & $\checkmark$ & & & & $57.869\pm4.65$ & $80.49\pm 4.85$ & $67.099\pm 3.075$\\
(c) & $\checkmark$ & $\checkmark$ & & $\checkmark$ & & & $62.569\pm7.059$ & $\textbf{89.45}\pm7.84$& $72.942\pm 2.644$\\
(d) & $\checkmark$& $\checkmark$ & $\checkmark$ &$\checkmark$ & & & $68.528\pm 9.412$ & $85.949\pm 11.3698$ & $74.934\pm 2.908$\\
(e) & $\checkmark$& $\checkmark$ & $\checkmark$ &$\checkmark$ & $\checkmark$ & & $69.444\pm 7.133$ & $86.638\pm 8.044$ & $76.385\pm 2.387$\\
\textbf{TFAD} & $\checkmark$ & $\checkmark$ & $\checkmark$ & $\checkmark$ & $\checkmark$ & $\checkmark$ & $\textbf{79.176}\pm 1.875$ & $85.231\pm 1.464$ & $\textbf{82.058}\pm 0.4199$\\
\hline
\end{tabular}
\end{center}
\end{table*}

For the multivariate time series anomaly detection in Table~\ref{tab:mulres}, we only compare TFAD with recent deep neural network models, since the conventional non-deep methods exhibit worse performance due to the limited ability for modeling the complex interaction and nonliterary of multivariate time series data. 
Note that all algorithms adopt unsupervised setting since none of multivariate datasets provides labels in the training data. 
It is interesting to find that our method even outperforms the state-of-the-art algorithms THOC~\cite{shen2020timeseries} and NCAD~\cite{carmona2021neural} by a reasonable margin. This is mainly due to our novel architecture with both time and frequency branches, while existing works detect anomalies only in the time domain.
In summary, our TFAD algorithm achieves the best F1 score among all competing algorithms in both SMAP and MSL datasets for multivariate time series anomaly detection.

% Most existing works detect anomalies only in the time domain. 
% Although with elaborate deep neural network design and the fact that most of the anomalies in existing public data sets are point-wise, OmniAnomaly, THOC, and so on gain good performances, few designs of sequence anomaly detection limit the scalability of the models. 

% % outperforms other methods in univariate and multivariate time series anomaly detection tasks. 
% As shown in Fig~\ref{network}, the anomaly score is computed. We can use such a score either in a supervised learning setting or in an unsupervised learning setting. 
% For the KPI data set, we considered both settings with/without giving the true data labels. 
% Note that, for unsupervised settings, although no true labels are used when training, labels are preliminary information for some data augmentation methods. 
% So we call this a semi-supervised method. 

% Actually, for multivariate time series, while we still use Hodrick-Prescott decomposition, which is designed for univariate time series, outstanding performances are gained. 
% A unique design for multivariate time series decomposition is future work.

\subsection{Ablation Studies}

To better understand how each component in TFAD contributes to the final accurate anomaly detection, we conduct ablation studies in the KPI dataset under supervised setting, and the results are summarized in Table~\ref{tab:ablation}. 

Firstly, when only time branch (base TCN model) or frequency branch (DFT followed by TCN) is adopted, it performs not well.
Secondly, with the decomposition module added in the time branch, the F1 score gains nearly 30\% improvement compared with the same base TCN model, which demonstrates the benefits of decomposition in TFAD model.
Thirdly, with the extra normal data augmentation module added where the ratio of augmentation data is set as 0.5, additional marginal improvement can be achieved. Similar improvements can be obtained with the time-domain abnormal data augmentation module added where the ratio of augmentation data is set as 0.4.
An interesting phenomenon is that, when both NormAug and TimeAnAug modules are added in the previous version, the improvement of the F1 score is more than the sum of improvement when they are added respectively. 
It can be explained that such two directions of data augmentation make the distance between normality and anomaly larger with a nearly multiplicative effect.
Note that the ratios of augmentation data may not be the best hyperparameters, but their performance improvements are still obvious. 
Lastly, after the frequency branch are added (corresponding to the full TFAD model), not only the F1 score improves, but the variance decreases. The reason is that, only with the time-domain branch without frequency branch, some sequence-wise anomalies are hard to be detected, and overfitting usually appears in the train set, which affects the performance in the test set. These ablation studies demonstrate the effectiveness of our TFAD design with time-frequency branches, data augmentation, and decomposition.

\subsection{Model Analysis and Discussion} 

% In this section, we analyze the performance of our model in theory and by visualization.

% % \noindent\textbf{Theory}

In this section, we provide visualizations and case studies to explain how the model works intuitively and obtain insights.

\subsubsection{Contribution of time series decomposition module}

\begin{figure}[t] 
\centering 
\subfigure[Original time series]{
\label{yahoowoD}
\includegraphics[width=0.23\textwidth]{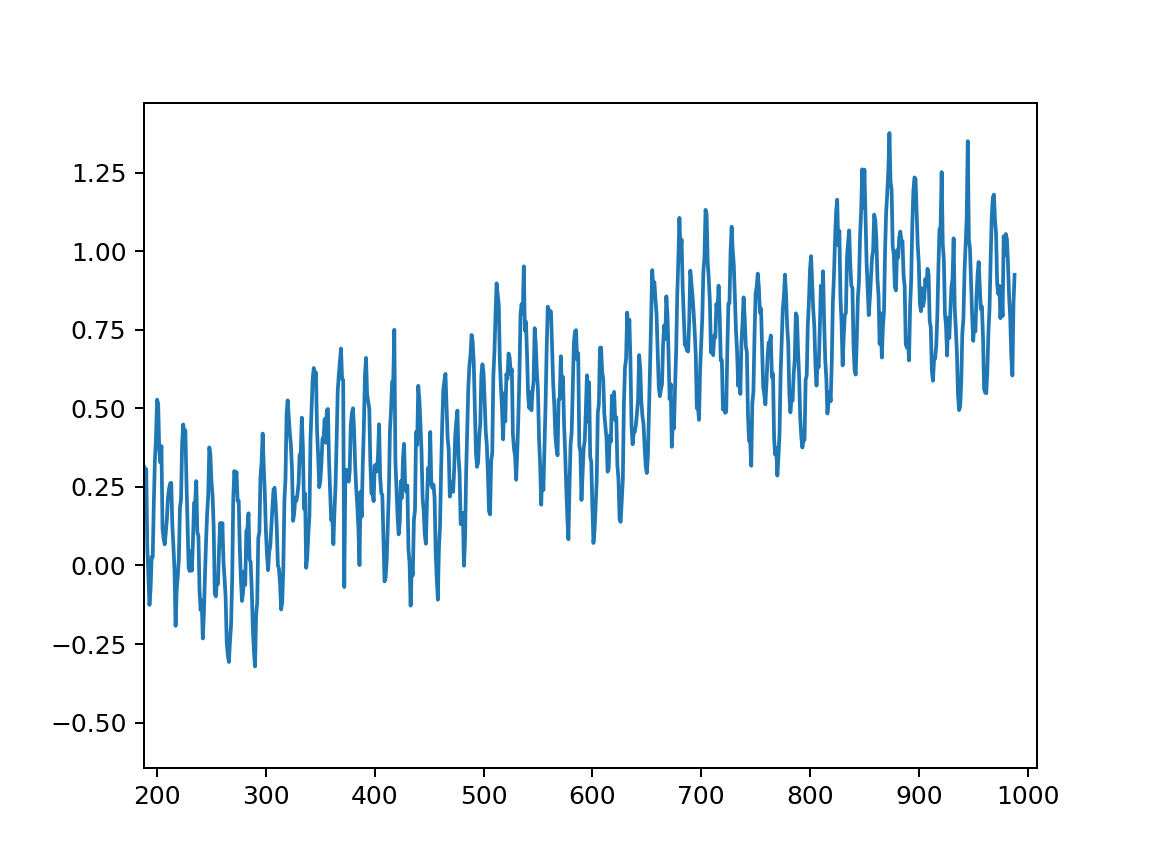}}
\subfigure[Time series with decomposition]{
\label{yahoowD}
\includegraphics[width=0.23\textwidth]{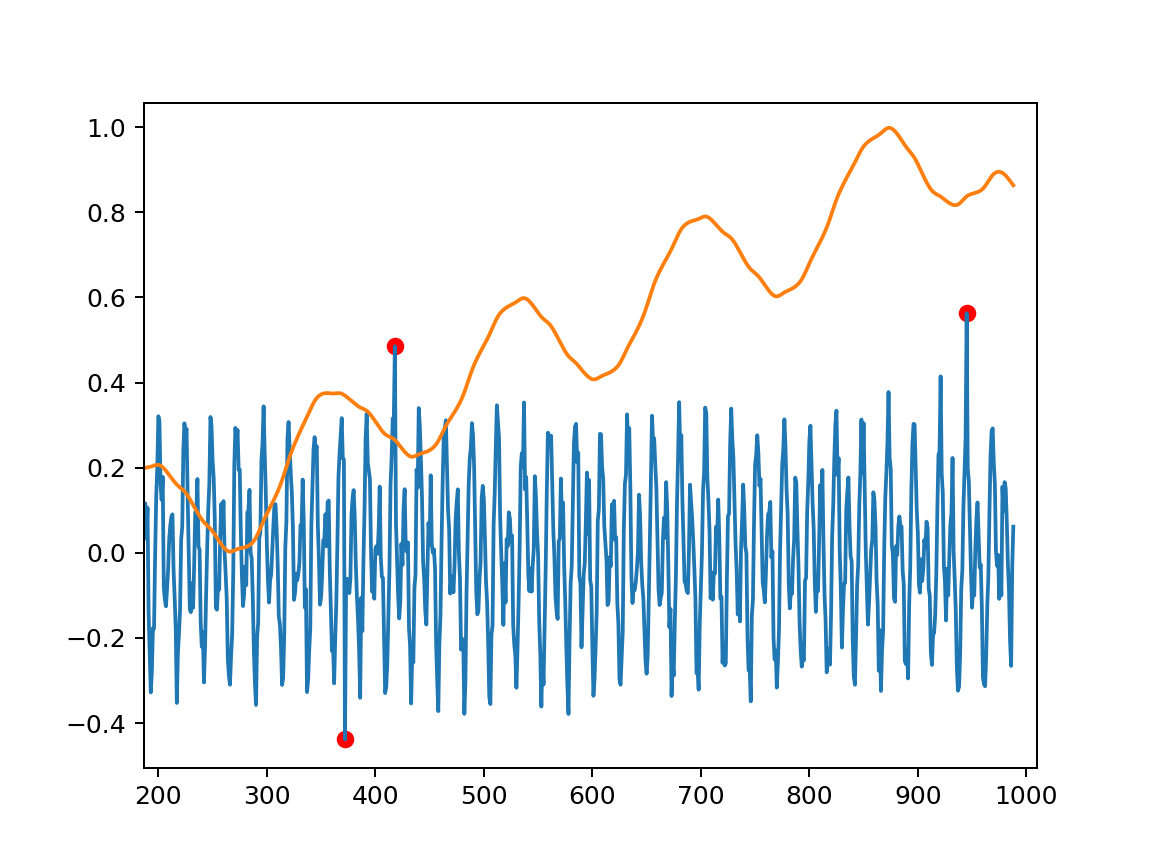}}
\caption{Examples on how decomposition helps time series anomaly detection.}
\label{yahooD} 
\end{figure}

Figure~\ref{yahooD} shows an example in the Yahoo data set to help understand how time series decomposition contributes to anomaly detection. Figure~\ref{yahoowoD} is the original time series without time series decomposition. Figure~\ref{yahoowD} shows the results of Figure~\ref{yahoowoD} after time series decomposition. The blue line is the residual component, the yellow line is the trend component, and the anomalies detected are labeled with red circles. Obviously, with the decomposition module, the anomalies are easier to detect. 

What is more, in our TFAD architecture as shown in Figure~\ref{network}, representation results can be gained for trend component and residual component independently, which makes it possible to obtain an anomaly score for each component and explain in which component anomaly happens.

\subsubsection{Effect of special anomaly data augmentation.}
Results in Table~\ref{tab:ablation} show that, with data augmentation added, performance can be improved. 
The anomaly data augmentation methods above are general and not designed for specific datasets. 
However, if prior information of the dataset is given, special anomaly data augmentation methods can be designed to take advantage of the pattern of anomalies for further performance improvements.

To demonstrate it, one case study is summarized in Table~\ref{tab:slow-slop} on SMAP dataset. The observation is that the first dimension of SMAP contains slow slops when anomalies appear. By utilizing this prior information, we conduct slow-slop injection on the first dimension of SMAP datasets as special anomaly data augmentation. With this specially designed data augmentation method, it can be seen in Table~\ref{tab:slow-slop} that significant extra performance gain is achieved in the TFAD model.

\begin {table}[h]
\caption {Results with slow-slop injection on first dimension of SMAP datasets as special anomaly data augmentation.} \label{tab:slow-slop}
\begin{center}
\begin{tabular}{c|c|c|c}
\hline
Model & Precision & Recall & F1 score \\
\hline
TFAD & 91.90 & 89.32 & 90.32\\
TFAD+injection & 94.04 & 98.36 & 96.09\\
\hline
\end{tabular}
\end{center}
\end{table}

% \begin{figure}[ht] 
% \centering 
% \includegraphics[scale=0.2]{KPIex2.png} 
% \caption{Example: Part of Anomalies Detected.} 
% \label{KPIEX2} 
% \end{figure}

%todo： trend anomaly, cycle anomaly 等找一些yahoo的结果，不过怎么说明这个比别的好呢？
% 统计一下KPI和Yahoo各自的段异常的比例？

\begin{figure}[!t]
\centering  
\subfigure[Typical anomalies detected]{
\label{KPIEX1}
\includegraphics[width=0.23\textwidth]{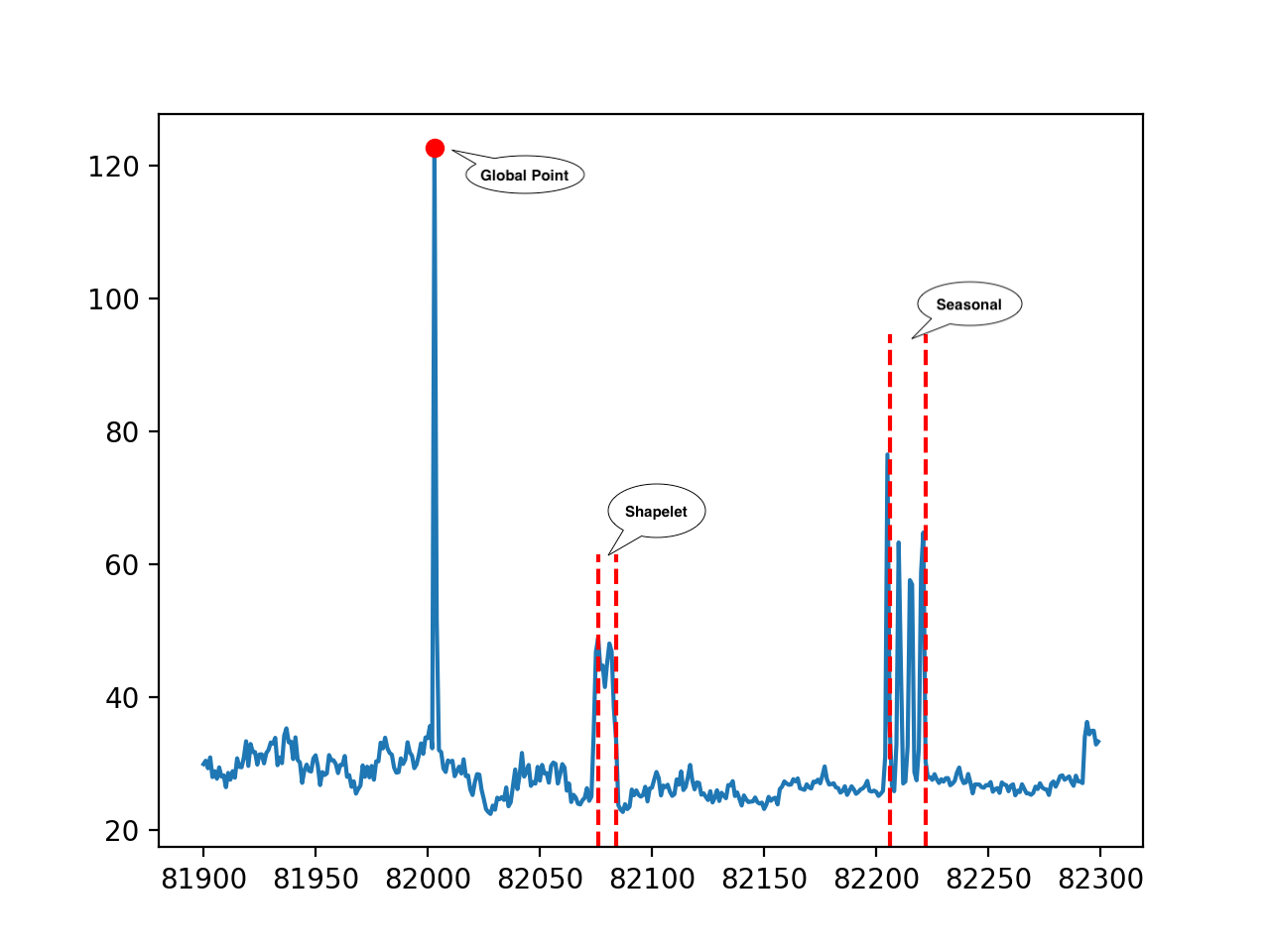}}
\subfigure[Part of anomalies detected]{
\label{KPIEX2}
\includegraphics[width=0.23\textwidth]{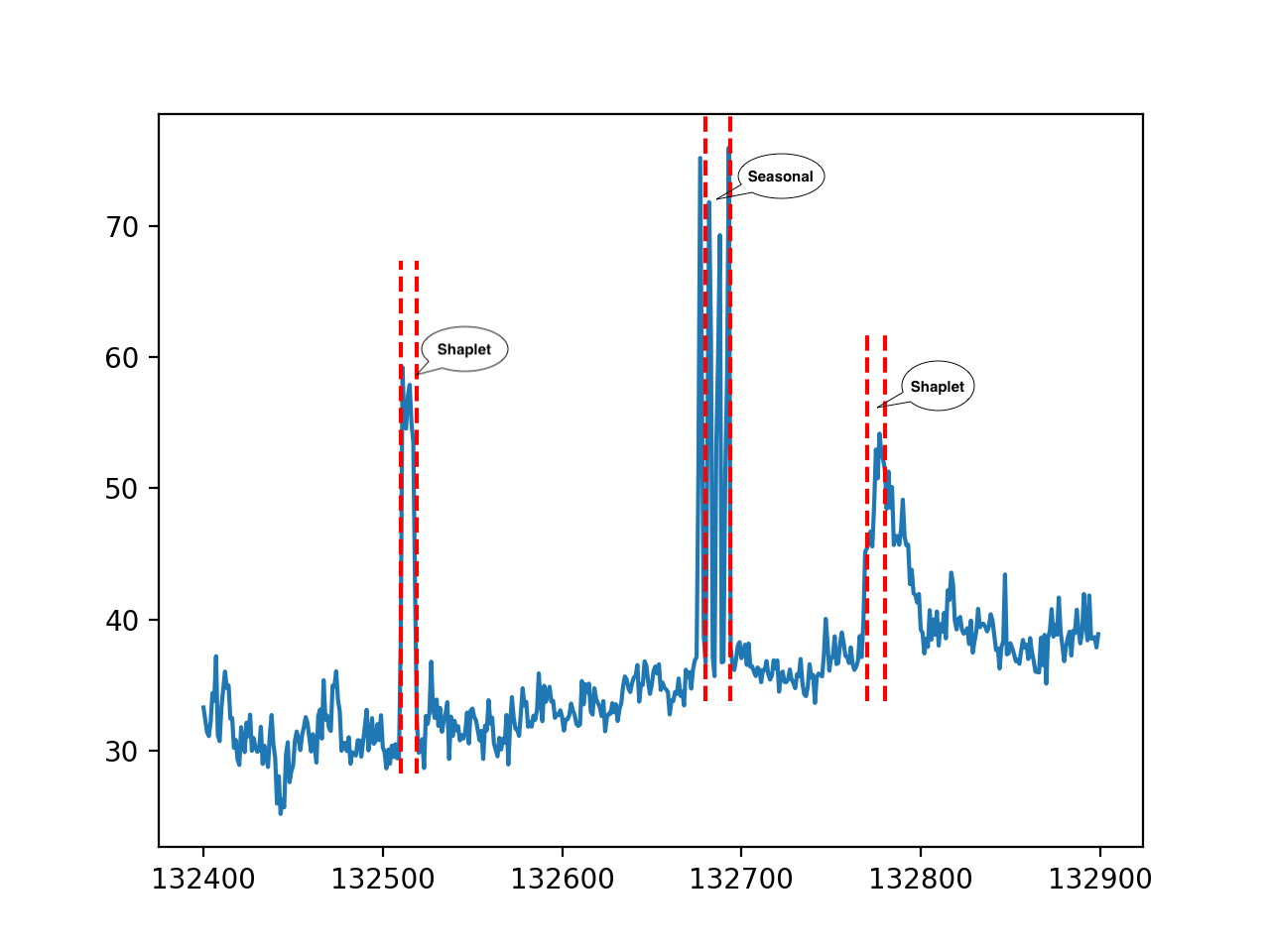}}
\caption{Detected anomaly examples of the proposed TFAD model on KPI datasets.}
\label{KPIEX}
\end{figure}
   
\subsubsection{Adaptive window size}
Several typical detected anomalies of the proposed TFAD model are shown in Figure~\ref{KPIEX}. In Figure~\ref{KPIEX1}, all the anomalies are detected, including point anomaly, shapelet anomaly, and seasonal anomaly. While in Figure~\ref{KPIEX2}, one interesting phenomenon is that part of the rightmost sub-sequence anomalies are not detected effectively. 
Considering that the anomalies in the middle and rightmost parts are near to each other and the anomalies in the middle part are more obvious than those in the rightmost part, thus when the suspect window is tested, the anomalies in the middle part change the representation of the full window and context window significantly, which will conceal the anomalies in the suspect window in the rightmost part. 
The adaptive window size may be a feasible solution for this problem, and we will leave it for future work. 
% The ratio and method of data augmentation also influence the final results significantly. It is in some way trifling to find the best ratios for different datasets. One promising solution is to choose them through adaptive learning. 
    
% 频域分支的方法：STFT等的效果对比

%%

\subsubsection{Representation Network}
In the TFAD model, simple temporal convolution neural (TCN) network is adopted as a representation network. 
We also tried the popular Transformer~\cite{vaswani2017attention} based representation learning networks for time series anomaly detection, since they are widely investigated in recent time series analysis~\cite{FedFormer,xu2021autoformer, zhou2021informer}. Unfortunately, besides the high memory usage of the Transformer, the F1 score is also much less than that of TCN: F1 score of 0.59 is gained with Transformer while F1 score of 0.75 is gained with TCN on KPI dataset. 
The reason is that the Transformer usually works well with long time series.
However, with window splitting for time series anomaly detection, the window length can not be set too long to distinguish between the full window and the context window. In this case, TCN can better model the local time series information in the window than Transformer networks.
Therefore, we adopt TCN network in the TFAD model, and it not only performs quite well in anomaly detection performance but also simplifies the implementation with low memory usage.

% since the decomposition in TFAD can simplify the complex time series. 
% In our setting, TCN is a good choice not only in respect of implementation but in performance. 

% the only difference between the full window and context window is the suspect window. The most significant need is to gain representation of the whole sequence of the full window and context window. 
% Hierarchical convolution network models the local information between near points better. 
% Therefore, it is essential in the time series anomaly detection task.

\section{Conclusion}\label{Conclusion}

This paper proposes a time-frequency analysis-based model (TFAD) for time series anomaly detection. 
Although most of the traditional and deep methods in time series anomaly detection have achieved great success, how to take advantage of the time-frequency properties of time series is not well investigated. Our design with time and frequency branches fills in this gap.
Besides, time series decomposition is implemented to bring insights into the explainability of the proposed model as well as simplify the neural network design. Furthermore, we also adopt data augmentation to overcome the lack of labeled anomaly data. 

Based on these considerations, the proposed TFAD with time-frequency architecture can handle the challenges of various anomalies in time series. Although no complex neural network architecture is implemented, we gain better performance than most existing deep models in time series anomaly detection. 
Extensive empirical studies with four benchmark datasets show that our TFAD scheme obtains a state-of-the-art performance. Furthermore, ablation studies show that TFAD gains not only higher accuracy but also lower variance for time series anomaly detection in both univariate and multivariate scenarios.

% We also discuss the explanation function of time series decomposition, the effect of specially designed anomaly injection methods, etc.
% A specially designed decomposition module for multi-variate time series and window size chosen with adaptive learning are left as future work. 
% we design a novel TFAD architecture. As it is hard to detect a deviation in time and frequency domain with single domain analysis, our two-branch design handles the challenges of various anomalies in time series.

% \vfill\pagebreak
% \clearpage

\bibliographystyle{ACM-Reference-Format.bst}
\bibliography{7_bibfile}

%%% -*-BibTeX-*-
%%% Do NOT edit. File created by BibTeX with style
%%% ACM-Reference-Format-Journals [18-Jan-2012].

\begin{thebibliography}{63}

%%% ====================================================================
%%% NOTE TO THE USER: you can override these defaults by providing
%%% customized versions of any of these macros before the \bibliography
%%% command.  Each of them MUST provide its own final punctuation,
%%% except for \shownote{}, \showDOI{}, and \showURL{}.  The latter two
%%% do not use final punctuation, in order to avoid confusing it with
%%% the Web address.
%%%
%%% To suppress output of a particular field, define its macro to expand
%%% to an empty string, or better, \unskip, like this:
%%%
%%% \newcommand{\showDOI}[1]{\unskip}   % LaTeX syntax
%%%
%%% \def \showDOI #1{\unskip}           % plain TeX syntax
%%%
%%% ====================================================================

\ifx \showCODEN    \undefined \def \showCODEN     #1{\unskip}     \fi
\ifx \showDOI      \undefined \def \showDOI       #1{#1}\fi
\ifx \showISBNx    \undefined \def \showISBNx     #1{\unskip}     \fi
\ifx \showISBNxiii \undefined \def \showISBNxiii  #1{\unskip}     \fi
\ifx \showISSN     \undefined \def \showISSN      #1{\unskip}     \fi
\ifx \showLCCN     \undefined \def \showLCCN      #1{\unskip}     \fi
\ifx \shownote     \undefined \def \shownote      #1{#1}          \fi
\ifx \showarticletitle \undefined \def \showarticletitle #1{#1}   \fi
\ifx \showURL      \undefined \def \showURL       {\relax}        \fi
% The following commands are used for tagged output and should be
% invisible to TeX
\providecommand\bibfield[2]{#2}
\providecommand\bibinfo[2]{#2}
\providecommand\natexlab[1]{#1}
\providecommand\showeprint[2][]{arXiv:#2}

\bibitem[Audibert et~al\mbox{.}(2020)]%
        {audibert2020usad}
\bibfield{author}{\bibinfo{person}{Julien Audibert}, \bibinfo{person}{Pietro
  Michiardi}, \bibinfo{person}{Fr{\'e}d{\'e}ric Guyard},
  \bibinfo{person}{S{\'e}bastien Marti}, {and} \bibinfo{person}{Maria~A
  Zuluaga}.} \bibinfo{year}{2020}\natexlab{}.
\newblock \showarticletitle{USAD: unsupervised anomaly detection on
  multivariate time series}. In \bibinfo{booktitle}{\emph{Proceedings of the
  26th ACM SIGKDD International Conference on Knowledge Discovery \& Data
  Mining}}. \bibinfo{pages}{3395--3404}.
\newblock


\bibitem[Bai et~al\mbox{.}(2018)]%
        {bai2018empirical}
\bibfield{author}{\bibinfo{person}{Shaojie Bai}, \bibinfo{person}{J~Zico
  Kolter}, {and} \bibinfo{person}{Vladlen Koltun}.}
  \bibinfo{year}{2018}\natexlab{}.
\newblock \showarticletitle{An empirical evaluation of generic convolutional
  and recurrent networks for sequence modeling}.
\newblock \bibinfo{journal}{\emph{arXiv preprint arXiv:1803.01271}}
  (\bibinfo{year}{2018}).
\newblock


\bibitem[Bl{\'a}zquez-Garc{\'\i}a et~al\mbox{.}(2021)]%
        {blazquez2021review}
\bibfield{author}{\bibinfo{person}{Ane Bl{\'a}zquez-Garc{\'\i}a},
  \bibinfo{person}{Angel Conde}, \bibinfo{person}{Usue Mori}, {and}
  \bibinfo{person}{Jose~A Lozano}.} \bibinfo{year}{2021}\natexlab{}.
\newblock \showarticletitle{A review on outlier/anomaly detection in time
  series data}.
\newblock \bibinfo{journal}{\emph{ACM Computing Surveys (CSUR)}}
  \bibinfo{volume}{54}, \bibinfo{number}{3} (\bibinfo{year}{2021}),
  \bibinfo{pages}{1--33}.
\newblock


\bibitem[Bock et~al\mbox{.}(2022)]%
        {bock2022online}
\bibfield{author}{\bibinfo{person}{Christian Bock},
  \bibinfo{person}{Fran{\c{c}}ois-Xavier Aubet}, \bibinfo{person}{Jan
  Gasthaus}, \bibinfo{person}{Andrey Kan}, \bibinfo{person}{Ming Chen}, {and}
  \bibinfo{person}{Laurent Callot}.} \bibinfo{year}{2022}\natexlab{}.
\newblock \showarticletitle{Online Time Series Anomaly Detection with State
  Space Gaussian Processes}.
\newblock \bibinfo{journal}{\emph{arXiv preprint arXiv:2201.06763}}
  (\bibinfo{year}{2022}).
\newblock


\bibitem[Carmona et~al\mbox{.}(2022)]%
        {carmona2021neural}
\bibfield{author}{\bibinfo{person}{Chris~U Carmona},
  \bibinfo{person}{Fran{\c{c}}ois-Xavier Aubet}, \bibinfo{person}{Valentin
  Flunkert}, {and} \bibinfo{person}{Jan Gasthaus}.}
  \bibinfo{year}{2022}\natexlab{}.
\newblock \showarticletitle{Neural contextual anomaly detection for time
  series}. In \bibinfo{booktitle}{\emph{Proceedings of International Joint
  Conference on Artificial Intelligence (IJCAI'22)}}.
\newblock


\bibitem[Chen et~al\mbox{.}(2022b)]%
        {quatformer22}
\bibfield{author}{\bibinfo{person}{Weiqi Chen}, \bibinfo{person}{Wenwei Wang},
  \bibinfo{person}{Bingqing Peng}, \bibinfo{person}{Qingsong Wen},
  \bibinfo{person}{Tian Zhou}, {and} \bibinfo{person}{Liang Sun}.}
  \bibinfo{year}{2022}\natexlab{b}.
\newblock \showarticletitle{Learning to Rotate: Quaternion Transformer for
  Complicated Periodical Time Series Forecasting}. In
  \bibinfo{booktitle}{\emph{Proceedings of the 28th ACM SIGKDD Conference on
  Knowledge Discovery and Data Mining}} (Washington DC, USA)
  \emph{(\bibinfo{series}{KDD '22})}. \bibinfo{pages}{146–156}.
\newblock


\bibitem[Chen et~al\mbox{.}(2022a)]%
        {chen2022adaptive}
\bibfield{author}{\bibinfo{person}{Zhuangbin Chen}, \bibinfo{person}{Jinyang
  Liu}, \bibinfo{person}{Yuxin Su}, \bibinfo{person}{Hongyu Zhang},
  \bibinfo{person}{Xiao Ling}, \bibinfo{person}{Yongqiang Yang}, {and}
  \bibinfo{person}{Michael~R Lyu}.} \bibinfo{year}{2022}\natexlab{a}.
\newblock \showarticletitle{Adaptive Performance Anomaly Detection for Online
  Service Systems via Pattern Sketching}.
\newblock \bibinfo{journal}{\emph{arXiv preprint arXiv:2201.02944}}
  (\bibinfo{year}{2022}).
\newblock


\bibitem[Cohen(1995)]%
        {cohen1995time}
\bibfield{author}{\bibinfo{person}{Leon Cohen}.}
  \bibinfo{year}{1995}\natexlab{}.
\newblock \bibinfo{booktitle}{\emph{Time-frequency analysis}}.
  Vol.~\bibinfo{volume}{778}.
\newblock \bibinfo{publisher}{Prentice hall New Jersey}.
\newblock


\bibitem[Competition(2018)]%
        {kpidata}
\bibfield{author}{\bibinfo{person}{AIOps Competition}.}
  \bibinfo{year}{2018}\natexlab{}.
\newblock \bibinfo{title}{KPI dataset}.
\newblock
\newblock
\urldef\tempurl%
\url{https://github.com/iopsai/iops/tree/master/phase2_env}
\showURL{%
\tempurl}


\bibitem[Dagum and Bianconcini(2016)]%
        {dagum2016seasonal}
\bibfield{author}{\bibinfo{person}{Estela~Bee Dagum} {and}
  \bibinfo{person}{Silvia Bianconcini}.} \bibinfo{year}{2016}\natexlab{}.
\newblock \bibinfo{booktitle}{\emph{Seasonal adjustment methods and real time
  trend-cycle estimation}}.
\newblock \bibinfo{publisher}{Springer}.
\newblock


\bibitem[Faloutsos et~al\mbox{.}(2020)]%
        {forecastFaloutsos20}
\bibfield{author}{\bibinfo{person}{Christos Faloutsos},
  \bibinfo{person}{Valentin Flunkert}, \bibinfo{person}{Jan Gasthaus},
  \bibinfo{person}{Tim Januschowski}, {and} \bibinfo{person}{Yuyang Wang}.}
  \bibinfo{year}{2020}\natexlab{}.
\newblock \showarticletitle{Forecasting Big Time Series: Theory and Practice}.
  In \bibinfo{booktitle}{\emph{Companion Proceedings of the Web Conference
  2020}} (Taipei, Taiwan) \emph{(\bibinfo{series}{WWW '20})}.
  \bibinfo{pages}{320–321}.
\newblock


\bibitem[Freeman et~al\mbox{.}(2021)]%
        {freeman2021experimental}
\bibfield{author}{\bibinfo{person}{Cynthia Freeman}, \bibinfo{person}{Jonathan
  Merriman}, \bibinfo{person}{Ian Beaver}, {and} \bibinfo{person}{Abdullah
  Mueen}.} \bibinfo{year}{2021}\natexlab{}.
\newblock \showarticletitle{Experimental Comparison and Survey of Twelve Time
  Series Anomaly Detection Algorithms}.
\newblock \bibinfo{journal}{\emph{Journal of Artificial Intelligence Research}}
   \bibinfo{volume}{72} (\bibinfo{year}{2021}), \bibinfo{pages}{849--899}.
\newblock


\bibitem[Gao et~al\mbox{.}(2002)]%
        {gao2002hmms}
\bibfield{author}{\bibinfo{person}{Bo Gao}, \bibinfo{person}{Hui-Ye Ma}, {and}
  \bibinfo{person}{Yu-Hang Yang}.} \bibinfo{year}{2002}\natexlab{}.
\newblock \showarticletitle{Hmms (hidden markov models) based on anomaly
  intrusion detection method}. In \bibinfo{booktitle}{\emph{Proceedings.
  International Conference on Machine Learning and Cybernetics}},
  Vol.~\bibinfo{volume}{1}. IEEE, \bibinfo{pages}{381--385}.
\newblock


\bibitem[Gao et~al\mbox{.}(2020)]%
        {gao2020robusttad}
\bibfield{author}{\bibinfo{person}{Jingkun Gao}, \bibinfo{person}{Xiaomin
  Song}, \bibinfo{person}{Qingsong Wen}, \bibinfo{person}{Pichao Wang},
  \bibinfo{person}{Liang Sun}, {and} \bibinfo{person}{Huan Xu}.}
  \bibinfo{year}{2020}\natexlab{}.
\newblock \showarticletitle{{RobustTAD}: Robust time series anomaly detection
  via decomposition and convolutional neural networks}.
\newblock \bibinfo{journal}{\emph{KDD Workshop on Mining and Learning from Time
  Series (KDD-MileTS'20)}} (\bibinfo{year}{2020}).
\newblock


\bibitem[Garg et~al\mbox{.}(2021)]%
        {garg2021evaluation}
\bibfield{author}{\bibinfo{person}{Astha Garg}, \bibinfo{person}{Wenyu Zhang},
  \bibinfo{person}{Jules Samaran}, \bibinfo{person}{Ramasamy Savitha}, {and}
  \bibinfo{person}{Chuan-Sheng Foo}.} \bibinfo{year}{2021}\natexlab{}.
\newblock \showarticletitle{An Evaluation of Anomaly Detection and Diagnosis in
  Multivariate Time Series}.
\newblock \bibinfo{journal}{\emph{IEEE Transactions on Neural Networks and
  Learning Systems}} (\bibinfo{year}{2021}).
\newblock


\bibitem[Guan et~al\mbox{.}(2016)]%
        {guan2016mapping}
\bibfield{author}{\bibinfo{person}{Xudong Guan}, \bibinfo{person}{Chong Huang},
  \bibinfo{person}{Gaohuan Liu}, \bibinfo{person}{Xuelian Meng}, {and}
  \bibinfo{person}{Qingsheng Liu}.} \bibinfo{year}{2016}\natexlab{}.
\newblock \showarticletitle{Mapping rice cropping systems in Vietnam using an
  NDVI-based time-series similarity measurement based on DTW distance}.
\newblock \bibinfo{journal}{\emph{Remote Sensing}} \bibinfo{volume}{8},
  \bibinfo{number}{1} (\bibinfo{year}{2016}), \bibinfo{pages}{19}.
\newblock


\bibitem[Gupta et~al\mbox{.}(2013)]%
        {gupta2013outlier}
\bibfield{author}{\bibinfo{person}{Manish Gupta}, \bibinfo{person}{Jing Gao},
  \bibinfo{person}{Charu~C Aggarwal}, {and} \bibinfo{person}{Jiawei Han}.}
  \bibinfo{year}{2013}\natexlab{}.
\newblock \showarticletitle{Outlier detection for temporal data: A survey}.
\newblock \bibinfo{journal}{\emph{IEEE Transactions on Knowledge and data
  Engineering}} \bibinfo{volume}{26}, \bibinfo{number}{9}
  (\bibinfo{year}{2013}), \bibinfo{pages}{2250--2267}.
\newblock


\bibitem[Hamilton(1994)]%
        {ts:textbook:1994}
\bibfield{author}{\bibinfo{person}{James~D. Hamilton}.}
  \bibinfo{year}{1994}\natexlab{}.
\newblock \bibinfo{booktitle}{\emph{{Time Series Analysis}}
  (\bibinfo{edition}{1} ed.)}.
\newblock \bibinfo{publisher}{Princeton University Press}.
\newblock
\showISBNx{0691042896}
\urldef\tempurl%
\url{http://www.amazon.com/exec/obidos/redirect?tag=citeulike07-20\&path=ASIN/0691042896}
\showURL{%
\tempurl}


\bibitem[Hendrycks et~al\mbox{.}(2018)]%
        {hendrycks2018deep}
\bibfield{author}{\bibinfo{person}{Dan Hendrycks}, \bibinfo{person}{Mantas
  Mazeika}, {and} \bibinfo{person}{Thomas Dietterich}.}
  \bibinfo{year}{2018}\natexlab{}.
\newblock \showarticletitle{Deep anomaly detection with outlier exposure}.
\newblock \bibinfo{journal}{\emph{arXiv preprint arXiv:1812.04606}}
  (\bibinfo{year}{2018}).
\newblock


\bibitem[Hochenbaum et~al\mbox{.}(2017)]%
        {hochenbaum2017automatic}
\bibfield{author}{\bibinfo{person}{Jordan Hochenbaum}, \bibinfo{person}{Owen~S
  Vallis}, {and} \bibinfo{person}{Arun Kejariwal}.}
  \bibinfo{year}{2017}\natexlab{}.
\newblock \showarticletitle{Automatic anomaly detection in the cloud via
  statistical learning}.
\newblock \bibinfo{journal}{\emph{arXiv preprint arXiv:1704.07706}}
  (\bibinfo{year}{2017}).
\newblock


\bibitem[Hodrick and Prescott(1997)]%
        {hodrick1997postwar}
\bibfield{author}{\bibinfo{person}{Robert~J Hodrick} {and}
  \bibinfo{person}{Edward~C Prescott}.} \bibinfo{year}{1997}\natexlab{}.
\newblock \showarticletitle{Postwar US business cycles: an empirical
  investigation}.
\newblock \bibinfo{journal}{\emph{Journal of Money, credit, and Banking}}
  (\bibinfo{year}{1997}), \bibinfo{pages}{1--16}.
\newblock


\bibitem[Hundman et~al\mbox{.}(2018)]%
        {hundman2018detecting}
\bibfield{author}{\bibinfo{person}{Kyle Hundman}, \bibinfo{person}{Valentino
  Constantinou}, \bibinfo{person}{Christopher Laporte}, \bibinfo{person}{Ian
  Colwell}, {and} \bibinfo{person}{Tom Soderstrom}.}
  \bibinfo{year}{2018}\natexlab{}.
\newblock \showarticletitle{Detecting spacecraft anomalies using lstms and
  nonparametric dynamic thresholding}. In \bibinfo{booktitle}{\emph{Proceedings
  of the 24th ACM SIGKDD international conference on knowledge discovery \&
  data mining}}. \bibinfo{pages}{387--395}.
\newblock


\bibitem[Hwang et~al\mbox{.}(2022)]%
        {hwang2022you}
\bibfield{author}{\bibinfo{person}{Won-Seok Hwang}, \bibinfo{person}{Jeong-Han
  Yun}, \bibinfo{person}{Jonguk Kim}, {and} \bibinfo{person}{Byung~Gil Min}.}
  \bibinfo{year}{2022}\natexlab{}.
\newblock \showarticletitle{Do you know existing accuracy metrics overrate
  time-series anomaly detections?}. In \bibinfo{booktitle}{\emph{Proceedings of
  the 37th ACM/SIGAPP Symposium on Applied Computing}}.
  \bibinfo{pages}{403--412}.
\newblock


\bibitem[Jacob et~al\mbox{.}(2020)]%
        {jacob2020exathlon}
\bibfield{author}{\bibinfo{person}{Vincent Jacob}, \bibinfo{person}{Fei Song},
  \bibinfo{person}{Arnaud Stiegler}, \bibinfo{person}{Bijan Rad},
  \bibinfo{person}{Yanlei Diao}, {and} \bibinfo{person}{Nesime Tatbul}.}
  \bibinfo{year}{2020}\natexlab{}.
\newblock \showarticletitle{Exathlon: a benchmark for explainable anomaly
  detection over time series}.
\newblock \bibinfo{journal}{\emph{arXiv preprint arXiv:2010.05073}}
  (\bibinfo{year}{2020}).
\newblock


\bibitem[Kim et~al\mbox{.}(2022)]%
        {kim2021towards}
\bibfield{author}{\bibinfo{person}{Siwon Kim}, \bibinfo{person}{Kukjin Choi},
  \bibinfo{person}{Hyun-Soo Choi}, \bibinfo{person}{Byunghan Lee}, {and}
  \bibinfo{person}{Sungroh Yoon}.} \bibinfo{year}{2022}\natexlab{}.
\newblock \showarticletitle{Towards a Rigorous Evaluation of Time-series
  Anomaly Detection}.
\newblock \bibinfo{journal}{\emph{AAAI}} (\bibinfo{year}{2022}).
\newblock


\bibitem[Kiran et~al\mbox{.}(2018)]%
        {kiran2018overview}
\bibfield{author}{\bibinfo{person}{B~Ravi Kiran}, \bibinfo{person}{Dilip~Mathew
  Thomas}, {and} \bibinfo{person}{Ranjith Parakkal}.}
  \bibinfo{year}{2018}\natexlab{}.
\newblock \showarticletitle{An overview of deep learning based methods for
  unsupervised and semi-supervised anomaly detection in videos}.
\newblock \bibinfo{journal}{\emph{Journal of Imaging}} \bibinfo{volume}{4},
  \bibinfo{number}{2} (\bibinfo{year}{2018}), \bibinfo{pages}{36}.
\newblock


\bibitem[Lai et~al\mbox{.}(2021)]%
        {lai2021revisiting}
\bibfield{author}{\bibinfo{person}{Kwei-Herng Lai}, \bibinfo{person}{Daochen
  Zha}, \bibinfo{person}{Junjie Xu}, \bibinfo{person}{Yue Zhao},
  \bibinfo{person}{Guanchu Wang}, {and} \bibinfo{person}{Xia Hu}.}
  \bibinfo{year}{2021}\natexlab{}.
\newblock \showarticletitle{Revisiting time series outlier detection:
  Definitions and benchmarks}. In \bibinfo{booktitle}{\emph{Thirty-fifth
  Conference on Neural Information Processing Systems Datasets and Benchmarks
  Track (Round 1)}}.
\newblock


\bibitem[Lane et~al\mbox{.}(1997)]%
        {lane1997sequence}
\bibfield{author}{\bibinfo{person}{Terran Lane}, \bibinfo{person}{Carla~E
  Brodley}, {et~al\mbox{.}}} \bibinfo{year}{1997}\natexlab{}.
\newblock \showarticletitle{Sequence matching and learning in anomaly detection
  for computer security}. In \bibinfo{booktitle}{\emph{AAAI Workshop: AI
  Approaches to Fraud Detection and Risk Management}}. Providence, Rhode
  Island, \bibinfo{pages}{43--49}.
\newblock


\bibitem[Li et~al\mbox{.}(2022)]%
        {li2022learning}
\bibfield{author}{\bibinfo{person}{Longyuan Li}, \bibinfo{person}{Junchi Yan},
  \bibinfo{person}{Qingsong Wen}, \bibinfo{person}{Yaohui Jin}, {and}
  \bibinfo{person}{Xiaokang Yang}.} \bibinfo{year}{2022}\natexlab{}.
\newblock \showarticletitle{Learning Robust Deep State Space for Unsupervised
  Anomaly Detection in Contaminated Time-Series}.
\newblock \bibinfo{journal}{\emph{IEEE TKDE}} (\bibinfo{year}{2022}).
\newblock


\bibitem[Muthukrishnan et~al\mbox{.}(2004)]%
        {muthukrishnan2004mining}
\bibfield{author}{\bibinfo{person}{Shan Muthukrishnan}, \bibinfo{person}{Rahul
  Shah}, {and} \bibinfo{person}{Jeffrey~Scott Vitter}.}
  \bibinfo{year}{2004}\natexlab{}.
\newblock \showarticletitle{Mining deviants in time series data streams}. In
  \bibinfo{booktitle}{\emph{Proceedings. 16th International Conference on
  Scientific and Statistical Database Management, 2004.}} IEEE,
  \bibinfo{pages}{41--50}.
\newblock


\bibitem[Parhizkar et~al\mbox{.}(2015)]%
        {parhizkar2015sequences}
\bibfield{author}{\bibinfo{person}{Reza Parhizkar}, \bibinfo{person}{Yann
  Barbotin}, {and} \bibinfo{person}{Martin Vetterli}.}
  \bibinfo{year}{2015}\natexlab{}.
\newblock \showarticletitle{Sequences with minimal time--frequency
  uncertainty}.
\newblock \bibinfo{journal}{\emph{Applied and Computational Harmonic Analysis}}
  \bibinfo{volume}{38}, \bibinfo{number}{3} (\bibinfo{year}{2015}),
  \bibinfo{pages}{452--468}.
\newblock


\bibitem[Park et~al\mbox{.}(2018)]%
        {park2018multimodal}
\bibfield{author}{\bibinfo{person}{Daehyung Park}, \bibinfo{person}{Yuuna
  Hoshi}, {and} \bibinfo{person}{Charles~C Kemp}.}
  \bibinfo{year}{2018}\natexlab{}.
\newblock \showarticletitle{A multimodal anomaly detector for robot-assisted
  feeding using an lstm-based variational autoencoder}.
\newblock \bibinfo{journal}{\emph{IEEE Robotics and Automation Letters}}
  \bibinfo{volume}{3}, \bibinfo{number}{3} (\bibinfo{year}{2018}),
  \bibinfo{pages}{1544--1551}.
\newblock


\bibitem[Patel et~al\mbox{.}(2022)]%
        {AnomalyKiTs_2022}
\bibfield{author}{\bibinfo{person}{Dhaval Patel}, \bibinfo{person}{Giridhar
  Ganapavarapu}, \bibinfo{person}{Srideepika Jayaraman},
  \bibinfo{person}{Shuxin Lin}, \bibinfo{person}{Anuradha Bhamidipaty}, {and}
  \bibinfo{person}{Jayant Kalagnanam}.} \bibinfo{year}{2022}\natexlab{}.
\newblock \showarticletitle{AnomalyKiTS: Anomaly Detection Toolkit for Time
  Series}.
\newblock \bibinfo{journal}{\emph{Proceedings of the AAAI Conference on
  Artificial Intelligence}} \bibinfo{volume}{36}, \bibinfo{number}{11}
  (\bibinfo{date}{Jun.} \bibinfo{year}{2022}), \bibinfo{pages}{13209--13211}.
\newblock


\bibitem[Ravn and Uhlig(2002)]%
        {ravn2002adjusting}
\bibfield{author}{\bibinfo{person}{Morten~O Ravn} {and} \bibinfo{person}{Harald
  Uhlig}.} \bibinfo{year}{2002}\natexlab{}.
\newblock \showarticletitle{On adjusting the Hodrick-Prescott filter for the
  frequency of observations}.
\newblock \bibinfo{journal}{\emph{Review of economics and statistics}}
  \bibinfo{volume}{84}, \bibinfo{number}{2} (\bibinfo{year}{2002}),
  \bibinfo{pages}{371--376}.
\newblock


\bibitem[Ren et~al\mbox{.}(2019)]%
        {ren2019time}
\bibfield{author}{\bibinfo{person}{Hansheng Ren}, \bibinfo{person}{Bixiong Xu},
  \bibinfo{person}{Yujing Wang}, \bibinfo{person}{Chao Yi},
  \bibinfo{person}{Congrui Huang}, \bibinfo{person}{Xiaoyu Kou},
  \bibinfo{person}{Tony Xing}, \bibinfo{person}{Mao Yang}, \bibinfo{person}{Jie
  Tong}, {and} \bibinfo{person}{Qi Zhang}.} \bibinfo{year}{2019}\natexlab{}.
\newblock \showarticletitle{Time-series anomaly detection service at
  microsoft}. In \bibinfo{booktitle}{\emph{Proceedings of the 25th ACM SIGKDD
  international conference on knowledge discovery \& data mining}}.
  \bibinfo{pages}{3009--3017}.
\newblock


\bibitem[Research(2015)]%
        {yahoolink}
\bibfield{author}{\bibinfo{person}{Yahoo Research}.}
  \bibinfo{year}{2015}\natexlab{}.
\newblock \showarticletitle{A Benchmark Dataset for Time Series Anomaly
  Detection,
  https://yahooresearch.tumblr.com/post/114590420346/a-benchmark-dataset-for-time-series-anomaly}.
\newblock  (\bibinfo{year}{2015}).
\newblock


\bibitem[Rezende et~al\mbox{.}(2014)]%
        {rezende2014stochastic}
\bibfield{author}{\bibinfo{person}{Danilo~Jimenez Rezende},
  \bibinfo{person}{Shakir Mohamed}, {and} \bibinfo{person}{Daan Wierstra}.}
  \bibinfo{year}{2014}\natexlab{}.
\newblock \showarticletitle{Stochastic backpropagation and approximate
  inference in deep generative models}. In
  \bibinfo{booktitle}{\emph{International conference on machine learning}}.
  PMLR, \bibinfo{pages}{1278--1286}.
\newblock


\bibitem[Robert et~al\mbox{.}(1990)]%
        {robert1990stl}
\bibfield{author}{\bibinfo{person}{Cleveland Robert}, \bibinfo{person}{C
  William}, {and} \bibinfo{person}{Terpenning Irma}.}
  \bibinfo{year}{1990}\natexlab{}.
\newblock \showarticletitle{STL: A seasonal-trend decomposition procedure based
  on loess}.
\newblock \bibinfo{journal}{\emph{Journal of official statistics}}
  \bibinfo{volume}{6}, \bibinfo{number}{1} (\bibinfo{year}{1990}),
  \bibinfo{pages}{3--73}.
\newblock


\bibitem[Ruff et~al\mbox{.}(2021)]%
        {ruff2021unifying}
\bibfield{author}{\bibinfo{person}{Lukas Ruff}, \bibinfo{person}{Jacob~R
  Kauffmann}, \bibinfo{person}{Robert~A Vandermeulen},
  \bibinfo{person}{Gr{\'e}goire Montavon}, \bibinfo{person}{Wojciech Samek},
  \bibinfo{person}{Marius Kloft}, \bibinfo{person}{Thomas~G Dietterich}, {and}
  \bibinfo{person}{Klaus-Robert M{\"u}ller}.} \bibinfo{year}{2021}\natexlab{}.
\newblock \showarticletitle{A unifying review of deep and shallow anomaly
  detection}.
\newblock \bibinfo{journal}{\emph{Proc. IEEE}} (\bibinfo{year}{2021}).
\newblock


\bibitem[Ruff et~al\mbox{.}(2018)]%
        {ruff2018deep}
\bibfield{author}{\bibinfo{person}{Lukas Ruff}, \bibinfo{person}{Robert
  Vandermeulen}, \bibinfo{person}{Nico Goernitz}, \bibinfo{person}{Lucas
  Deecke}, \bibinfo{person}{Shoaib~Ahmed Siddiqui}, \bibinfo{person}{Alexander
  Binder}, \bibinfo{person}{Emmanuel M{\"u}ller}, {and} \bibinfo{person}{Marius
  Kloft}.} \bibinfo{year}{2018}\natexlab{}.
\newblock \showarticletitle{Deep one-class classification}. In
  \bibinfo{booktitle}{\emph{International conference on machine learning}}.
  PMLR, \bibinfo{pages}{4393--4402}.
\newblock


\bibitem[Ruff et~al\mbox{.}(2020)]%
        {ruff2020rethinking}
\bibfield{author}{\bibinfo{person}{Lukas Ruff}, \bibinfo{person}{Robert~A
  Vandermeulen}, \bibinfo{person}{Billy~Joe Franks},
  \bibinfo{person}{Klaus-Robert M{\"u}ller}, {and} \bibinfo{person}{Marius
  Kloft}.} \bibinfo{year}{2020}\natexlab{}.
\newblock \showarticletitle{Rethinking assumptions in deep anomaly detection}.
\newblock \bibinfo{journal}{\emph{arXiv preprint arXiv:2006.00339}}
  (\bibinfo{year}{2020}).
\newblock


\bibitem[Schlegl et~al\mbox{.}(2017)]%
        {schlegl2017unsupervised}
\bibfield{author}{\bibinfo{person}{Thomas Schlegl}, \bibinfo{person}{Philipp
  Seeb{\"o}ck}, \bibinfo{person}{Sebastian~M Waldstein},
  \bibinfo{person}{Ursula Schmidt-Erfurth}, {and} \bibinfo{person}{Georg
  Langs}.} \bibinfo{year}{2017}\natexlab{}.
\newblock \showarticletitle{Unsupervised anomaly detection with generative
  adversarial networks to guide marker discovery}. In
  \bibinfo{booktitle}{\emph{International conference on information processing
  in medical imaging}}. Springer, \bibinfo{pages}{146--157}.
\newblock


\bibitem[Shah et~al\mbox{.}(2021)]%
        {AutoAI-TS}
\bibfield{author}{\bibinfo{person}{Syed~Yousaf Shah}, \bibinfo{person}{Dhaval
  Patel}, \bibinfo{person}{Long Vu}, \bibinfo{person}{Xuan-Hong Dang},
  \bibinfo{person}{Bei Chen}, \bibinfo{person}{Peter Kirchner},
  \bibinfo{person}{Horst Samulowitz}, \bibinfo{person}{David Wood},
  \bibinfo{person}{Gregory Bramble}, \bibinfo{person}{Wesley~M. Gifford},
  \bibinfo{person}{Giridhar Ganapavarapu}, \bibinfo{person}{Roman Vaculin},
  {and} \bibinfo{person}{Petros Zerfos}.} \bibinfo{year}{2021}\natexlab{}.
\newblock \showarticletitle{AutoAI-TS: AutoAI for Time Series Forecasting}. In
  \bibinfo{booktitle}{\emph{Proceedings of the 2021 International Conference on
  Management of Data}} (Virtual Event, China) \emph{(\bibinfo{series}{SIGMOD
  '21})}. \bibinfo{pages}{2584–2596}.
\newblock


\bibitem[Shen et~al\mbox{.}(2020)]%
        {shen2020timeseries}
\bibfield{author}{\bibinfo{person}{Lifeng Shen}, \bibinfo{person}{Zhuocong Li},
  {and} \bibinfo{person}{James Kwok}.} \bibinfo{year}{2020}\natexlab{}.
\newblock \showarticletitle{Timeseries anomaly detection using temporal
  hierarchical one-class network}.
\newblock \bibinfo{journal}{\emph{Advances in Neural Information Processing
  Systems}}  \bibinfo{volume}{33} (\bibinfo{year}{2020}),
  \bibinfo{pages}{13016--13026}.
\newblock


\bibitem[Shorten and Khoshgoftaar(2019)]%
        {shorten2019survey}
\bibfield{author}{\bibinfo{person}{Connor Shorten} {and}
  \bibinfo{person}{Taghi~M Khoshgoftaar}.} \bibinfo{year}{2019}\natexlab{}.
\newblock \showarticletitle{A survey on image data augmentation for deep
  learning}.
\newblock \bibinfo{journal}{\emph{Journal of big data}} \bibinfo{volume}{6},
  \bibinfo{number}{1} (\bibinfo{year}{2019}), \bibinfo{pages}{1--48}.
\newblock


\bibitem[Shumway et~al\mbox{.}(2000)]%
        {shumway2000time}
\bibfield{author}{\bibinfo{person}{Robert~H Shumway}, \bibinfo{person}{David~S
  Stoffer}, {and} \bibinfo{person}{David~S Stoffer}.}
  \bibinfo{year}{2000}\natexlab{}.
\newblock \bibinfo{booktitle}{\emph{Time series analysis and its
  applications}}. Vol.~\bibinfo{volume}{3}.
\newblock \bibinfo{publisher}{Springer}.
\newblock


\bibitem[Siffer et~al\mbox{.}(2017)]%
        {siffer2017anomaly}
\bibfield{author}{\bibinfo{person}{Alban Siffer}, \bibinfo{person}{Pierre-Alain
  Fouque}, \bibinfo{person}{Alexandre Termier}, {and}
  \bibinfo{person}{Christine Largouet}.} \bibinfo{year}{2017}\natexlab{}.
\newblock \showarticletitle{Anomaly detection in streams with extreme value
  theory}. In \bibinfo{booktitle}{\emph{Proceedings of the 23rd ACM SIGKDD
  International Conference on Knowledge Discovery and Data Mining}}.
  \bibinfo{pages}{1067--1075}.
\newblock


\bibitem[Su et~al\mbox{.}(2019)]%
        {su2019robust}
\bibfield{author}{\bibinfo{person}{Ya Su}, \bibinfo{person}{Youjian Zhao},
  \bibinfo{person}{Chenhao Niu}, \bibinfo{person}{Rong Liu},
  \bibinfo{person}{Wei Sun}, {and} \bibinfo{person}{Dan Pei}.}
  \bibinfo{year}{2019}\natexlab{}.
\newblock \showarticletitle{Robust anomaly detection for multivariate time
  series through stochastic recurrent neural network}. In
  \bibinfo{booktitle}{\emph{Proceedings of the 25th ACM SIGKDD international
  conference on knowledge discovery \& data mining}}.
  \bibinfo{pages}{2828--2837}.
\newblock


\bibitem[Tuli et~al\mbox{.}(2022)]%
        {tuli2022tranad}
\bibfield{author}{\bibinfo{person}{Shreshth Tuli}, \bibinfo{person}{Giuliano
  Casale}, {and} \bibinfo{person}{Nicholas~R Jennings}.}
  \bibinfo{year}{2022}\natexlab{}.
\newblock \showarticletitle{TranAD: Deep Transformer Networks for Anomaly
  Detection in Multivariate Time Series Data}. In
  \bibinfo{booktitle}{\emph{Proceedings of 48th International Conference on
  Very Large Databases (VLDB'22)}}.
\newblock


\bibitem[Vaswani et~al\mbox{.}(2017)]%
        {vaswani2017attention}
\bibfield{author}{\bibinfo{person}{Ashish Vaswani}, \bibinfo{person}{Noam
  Shazeer}, \bibinfo{person}{Niki Parmar}, \bibinfo{person}{Jakob Uszkoreit},
  \bibinfo{person}{Llion Jones}, \bibinfo{person}{Aidan~N Gomez},
  \bibinfo{person}{{\L}ukasz Kaiser}, {and} \bibinfo{person}{Illia
  Polosukhin}.} \bibinfo{year}{2017}\natexlab{}.
\newblock \showarticletitle{Attention is all you need}.
\newblock \bibinfo{journal}{\emph{Advances in neural information processing
  systems}}  \bibinfo{volume}{30} (\bibinfo{year}{2017}).
\newblock


\bibitem[Wen et~al\mbox{.}(2019)]%
        {wen2019robuststl}
\bibfield{author}{\bibinfo{person}{Qingsong Wen}, \bibinfo{person}{Jingkun
  Gao}, \bibinfo{person}{Xiaomin Song}, \bibinfo{person}{Liang Sun},
  \bibinfo{person}{Huan Xu}, {and} \bibinfo{person}{Shenghuo Zhu}.}
  \bibinfo{year}{2019}\natexlab{}.
\newblock \showarticletitle{{RobustSTL}: A robust seasonal-trend decomposition
  algorithm for long time series}. In \bibinfo{booktitle}{\emph{Proceedings of
  the AAAI Conference on Artificial Intelligence (AAAI'19)}},
  Vol.~\bibinfo{volume}{33}. \bibinfo{pages}{5409--5416}.
\newblock


\bibitem[Wen et~al\mbox{.}(2021a)]%
        {robustPeriod:2021}
\bibfield{author}{\bibinfo{person}{Qingsong Wen}, \bibinfo{person}{Kai He},
  \bibinfo{person}{Liang Sun}, \bibinfo{person}{Yingying Zhang},
  \bibinfo{person}{Min Ke}, {and} \bibinfo{person}{Huan Xu}.}
  \bibinfo{year}{2021}\natexlab{a}.
\newblock \showarticletitle{RobustPeriod: Robust Time-Frequency Mining for
  Multiple Periodicity Detection}. In \bibinfo{booktitle}{\emph{Proceedings of
  the 2021 International Conference on Management of Data (SIGMOD'21)}}.
  \bibinfo{pages}{2328–2337}.
\newblock


\bibitem[Wen et~al\mbox{.}(2021b)]%
        {wen2020time}
\bibfield{author}{\bibinfo{person}{Qingsong Wen}, \bibinfo{person}{Liang Sun},
  \bibinfo{person}{Fan Yang}, \bibinfo{person}{Xiaomin Song},
  \bibinfo{person}{Jingkun Gao}, \bibinfo{person}{Xue Wang}, {and}
  \bibinfo{person}{Huan Xu}.} \bibinfo{year}{2021}\natexlab{b}.
\newblock \showarticletitle{Time Series Data Augmentation for Deep Learning: A
  Survey}. In \bibinfo{booktitle}{\emph{Proceedings of International Joint
  Conference on Artificial Intelligence (IJCAI'21)}}.
  \bibinfo{pages}{4653--4660}.
\newblock


\bibitem[Wen et~al\mbox{.}(2022)]%
        {robustts22}
\bibfield{author}{\bibinfo{person}{Qingsong Wen}, \bibinfo{person}{Linxiao
  Yang}, \bibinfo{person}{Tian Zhou}, {and} \bibinfo{person}{Liang Sun}.}
  \bibinfo{year}{2022}\natexlab{}.
\newblock \showarticletitle{Robust Time Series Analysis and Applications: An
  Industrial Perspective}. In \bibinfo{booktitle}{\emph{Proceedings of the 28th
  ACM SIGKDD Conference on Knowledge Discovery and Data Mining}} (Washington
  DC, USA) \emph{(\bibinfo{series}{KDD '22})}. \bibinfo{pages}{4836–4837}.
\newblock


\bibitem[Xu et~al\mbox{.}(2018)]%
        {xu2018unsupervised}
\bibfield{author}{\bibinfo{person}{Haowen Xu}, \bibinfo{person}{Wenxiao Chen},
  \bibinfo{person}{Nengwen Zhao}, \bibinfo{person}{Zeyan Li},
  \bibinfo{person}{Jiahao Bu}, \bibinfo{person}{Zhihan Li},
  \bibinfo{person}{Ying Liu}, \bibinfo{person}{Youjian Zhao},
  \bibinfo{person}{Dan Pei}, \bibinfo{person}{Yang Feng}, {et~al\mbox{.}}}
  \bibinfo{year}{2018}\natexlab{}.
\newblock \showarticletitle{Unsupervised anomaly detection via variational
  auto-encoder for seasonal kpis in web applications}. In
  \bibinfo{booktitle}{\emph{Proceedings of the 2018 world wide web
  conference}}. \bibinfo{pages}{187--196}.
\newblock


\bibitem[Xu et~al\mbox{.}(2021)]%
        {xu2021autoformer}
\bibfield{author}{\bibinfo{person}{Jiehui Xu}, \bibinfo{person}{Jianmin Wang},
  \bibinfo{person}{Mingsheng Long}, {et~al\mbox{.}}}
  \bibinfo{year}{2021}\natexlab{}.
\newblock \showarticletitle{Autoformer: Decomposition transformers with
  auto-correlation for long-term series forecasting}.
\newblock \bibinfo{journal}{\emph{Advances in Neural Information Processing
  Systems}}  \bibinfo{volume}{34} (\bibinfo{year}{2021}).
\newblock


\bibitem[Xu et~al\mbox{.}(2022)]%
        {xu2021anomaly}
\bibfield{author}{\bibinfo{person}{Jiehui Xu}, \bibinfo{person}{Haixu Wu},
  \bibinfo{person}{Jianmin Wang}, {and} \bibinfo{person}{Mingsheng Long}.}
  \bibinfo{year}{2022}\natexlab{}.
\newblock \showarticletitle{Anomaly Transformer: Time Series Anomaly Detection
  with Association Discrepancy}. In \bibinfo{booktitle}{\emph{International
  Conference on Learning Representations}}.
\newblock


\bibitem[Zhang et~al\mbox{.}(2019)]%
        {zhang2019deep}
\bibfield{author}{\bibinfo{person}{Chuxu Zhang}, \bibinfo{person}{Dongjin
  Song}, \bibinfo{person}{Yuncong Chen}, \bibinfo{person}{Xinyang Feng},
  \bibinfo{person}{Cristian Lumezanu}, \bibinfo{person}{Wei Cheng},
  \bibinfo{person}{Jingchao Ni}, \bibinfo{person}{Bo Zong},
  \bibinfo{person}{Haifeng Chen}, {and} \bibinfo{person}{Nitesh~V Chawla}.}
  \bibinfo{year}{2019}\natexlab{}.
\newblock \showarticletitle{A deep neural network for unsupervised anomaly
  detection and diagnosis in multivariate time series data}. In
  \bibinfo{booktitle}{\emph{Proceedings of the AAAI conference on artificial
  intelligence (AAAI'19)}}, Vol.~\bibinfo{volume}{33}.
  \bibinfo{pages}{1409--1416}.
\newblock


\bibitem[Zhang et~al\mbox{.}(2021)]%
        {zhang2021cloudrca}
\bibfield{author}{\bibinfo{person}{Yingying Zhang}, \bibinfo{person}{Zhengxiong
  Guan}, \bibinfo{person}{Huajie Qian}, \bibinfo{person}{Leili Xu},
  \bibinfo{person}{Hengbo Liu}, \bibinfo{person}{Qingsong Wen},
  \bibinfo{person}{Liang Sun}, \bibinfo{person}{Junwei Jiang},
  \bibinfo{person}{Lunting Fan}, {and} \bibinfo{person}{Min Ke}.}
  \bibinfo{year}{2021}\natexlab{}.
\newblock \showarticletitle{{CloudRCA}: A Root Cause Analysis Framework for
  Cloud Computing Platforms}. In \bibinfo{booktitle}{\emph{CIKM 2021}}.
\newblock


\bibitem[Zhao et~al\mbox{.}(2020)]%
        {zhao2020multivariate}
\bibfield{author}{\bibinfo{person}{Hang Zhao}, \bibinfo{person}{Yujing Wang},
  \bibinfo{person}{Juanyong Duan}, \bibinfo{person}{Congrui Huang},
  \bibinfo{person}{Defu Cao}, \bibinfo{person}{Yunhai Tong},
  \bibinfo{person}{Bixiong Xu}, \bibinfo{person}{Jing Bai},
  \bibinfo{person}{Jie Tong}, {and} \bibinfo{person}{Qi Zhang}.}
  \bibinfo{year}{2020}\natexlab{}.
\newblock \showarticletitle{Multivariate time-series anomaly detection via
  graph attention network}. In \bibinfo{booktitle}{\emph{2020 IEEE
  International Conference on Data Mining (ICDM)}}. IEEE,
  \bibinfo{pages}{841--850}.
\newblock


\bibitem[Zhou et~al\mbox{.}(2021)]%
        {zhou2021informer}
\bibfield{author}{\bibinfo{person}{Haoyi Zhou}, \bibinfo{person}{Shanghang
  Zhang}, \bibinfo{person}{Jieqi Peng}, \bibinfo{person}{Shuai Zhang},
  \bibinfo{person}{Jianxin Li}, \bibinfo{person}{Hui Xiong}, {and}
  \bibinfo{person}{Wancai Zhang}.} \bibinfo{year}{2021}\natexlab{}.
\newblock \showarticletitle{Informer: Beyond efficient transformer for long
  sequence time-series forecasting}. In \bibinfo{booktitle}{\emph{Proceedings
  of the AAAI Conference on Artificial Intelligence (AAAI'21)}}.
\newblock


\bibitem[Zhou et~al\mbox{.}(2022)]%
        {FedFormer}
\bibfield{author}{\bibinfo{person}{Tian Zhou}, \bibinfo{person}{Ziqing Ma},
  \bibinfo{person}{Qingsong Wen}, \bibinfo{person}{Xue Wang},
  \bibinfo{person}{Liang Sun}, {and} \bibinfo{person}{Rong Jin}.}
  \bibinfo{year}{2022}\natexlab{}.
\newblock \showarticletitle{{FEDformer}: Frequency enhanced decomposed
  transformer for long-term series forecasting}. In
  \bibinfo{booktitle}{\emph{39th International Conference on Machine Learning
  (ICML)}}.
\newblock


\bibitem[Zong et~al\mbox{.}(2018)]%
        {zong2018deep}
\bibfield{author}{\bibinfo{person}{Bo Zong}, \bibinfo{person}{Qi Song},
  \bibinfo{person}{Martin~Renqiang Min}, \bibinfo{person}{Wei Cheng},
  \bibinfo{person}{Cristian Lumezanu}, \bibinfo{person}{Daeki Cho}, {and}
  \bibinfo{person}{Haifeng Chen}.} \bibinfo{year}{2018}\natexlab{}.
\newblock \showarticletitle{Deep autoencoding gaussian mixture model for
  unsupervised anomaly detection}. In \bibinfo{booktitle}{\emph{International
  conference on learning representations (ICLR'18)}}.
\newblock


\end{thebibliography}

\end{document}